\documentclass[sigconf]{aamas}

\usepackage{balance} 

\usepackage{subfigure}
\usepackage{glossaries}
\usepackage{multirow}
\usepackage{hyperref}
\usepackage[vlined,ruled,linesnumbered]{algorithm2e}

\setcopyright{ifaamas}
\acmConference[AAMAS '25]{Proc.\@ of the 24th International Conference
on Autonomous Agents and Multiagent Systems (AAMAS 2025)}{May 19 -- 23, 2025}
{Detroit, Michigan, USA}{A.~El~Fallah~Seghrouchni, Y.~Vorobeychik, S.~Das, A.~Nowe (eds.)}
\copyrightyear{2025}
\acmYear{2025}
\acmDOI{}
\acmPrice{}
\acmISBN{}

\acmSubmissionID{155}

\title[AAMAS-2025 Formatting Instructions]{Transformer Guided Coevolution: Improved Team Selection in Multiagent Adversarial Team Games}

\author{Pranav Rajbhandari}
\affiliation{
  \institution{Carnegie Mellon University}
  \city{Pittsburgh, PA}
  \country{United States}}
\email{prajbhan@alumni.cmu.edu}

\author{Prithviraj Dasgupta}
\affiliation{
  \institution{Naval Research Laboratory}
  \city{Washington, D.C.}
  \country{United States}}
\email{prithviraj.dasgupta.civ@us.navy.mil}

\author{Donald Sofge}
\affiliation{
  \institution{Naval Research Laboratory}
  \city{Washington, D.C.}
  \country{United States}}
\email{donald.a.sofge.civ@us.navy.mil}

\newcommand{\todo}[1]{}

\newcommand{\lrquote}[1]{\lq#1\rq}
\newcommand{\commentout}[1]{}

\def\halffiguresize{.49\linewidth}

\numberwithin{define}{section}

\newacronym{RL}{RL}{Reinforcement Learning}
\newacronym{QD}{QD}{Quality Diversity}
\newacronym{MCAA}{MCAA}{Multiagent Coevolution for Asymmetric Agents}
\newacronym{MDP}{MDP}{Markov Decision Process}
\newacronym{NLP}{NLP}{Natural Language Processing}
\newacronym{PPO}{PPO}{Proximal Policy Optimization}
\newacronym{MLM}{MLM}{Masked Language Modeling}
\newacronym{BERT}{BERT}{Bidirectional Encoder Representations from Transformers}
\begin{abstract}
We consider the problem of team selection within multiagent adversarial team games. 
We propose BERTeam, a novel algorithm that uses a transformer-based deep neural network with Masked Language Model training to select the best team of players from a trained population.
We integrate this with coevolutionary deep reinforcement learning, which trains a diverse set of individual players to choose from. 
We test our algorithm in the multiagent adversarial game Marine Capture-The-Flag, and find that BERTeam learns non-trivial team compositions that perform well against unseen opponents. 
For this game, we find that BERTeam outperforms MCAA, an algorithm that similarly optimizes team selection.
\end{abstract}


\newcommand{\BibTeX}{\rm B\kern-.05em{\sc i\kern-.025em b}\kern-.08em\TeX}

\begin{document}


\pagestyle{fancy}
\fancyhead{}


\maketitle 


%

\section{Introduction}

We inspect multiagent adversarial team games, characterized by an environment with multiple teams of agents, each working to achieve a team goal. 
Their performance is evaluated by an outcome, a real number assigned to each team at the end of a game (episode).

Various complex team games can be formulated as a multiagent adversarial team game, including pursuit-evasion games \cite{reachavoid2, reachavoid1, team_comp_domain_specific, teamformationassesor}, robotic football \cite{robosoccer,robocup,googlefootball}, and robotic capture-the-flag \cite{aquaticus}. 
The problem of creating a cooperative team of robots is useful for solving these games, as well as applications like search and rescue.

A crucial problem in these games is the selection of teams from a set of potential members to perform well against unknown opponents.
This problem is difficult since a team selection algorithm must consider both intra-team and inter-team interactions to select an optimal team.
Additionally, the set of agents often must learn their individual policies, increasing the complexity of the problem.

Researchers have addressed the team selection problem using evolutionary computation-based approaches \cite{Dixit22,mapelite}, albeit for non-adversarial settings like search and reconnaissance. 
In this paper, we consider the use of a transformer based neural network to predict a set of agents to form a team.
We name this technique BERTeam, and investigate its suitability for team selection in multiagent adversarial team games.

In contrast to evolution-based approaches in literature, our technique considers team selection as a token sequence generation process. 
We generate a team by beginning with a masked sequence of members, and iteratively querying the transformer to predict the next masked agent's identity.
We continue until the specified team size is reached.
We train BERTeam on a dataset of well-performing teams automatically generated from outcomes of sampled games.

Alongside training BERTeam, we evolve a population of agents using Coevolutionary Deep Reinforcement Learning \cite{coevdeeprl, coevdeeprl2}. 
This method utilizes self-play, sampling games between teams selected from the population. The training data from these games updates a set of \gls{RL} algorithms and guides a standard evolutionary algorithm.

We empirically validate our proposed technique with the $k$-v-$k$ adversarial game Marine Capture-the-Flag (MCTF). 
Our results show that BERTeam is an effective method for team selection in this domain.
We found that BERTeam is able to learn a non-trivial distribution, favoring well performing teams.
We also find that for MCTF, BERTeam outperforms \gls{MCAA}, another team selection algorithm.

\commentout{
\subsection{Document Structure}
\todo{maybe remove this}
In this section, we introduce multiagent adversarial team games and summarize our goal.
We describe background information and related work in Sections \ref{background} and \ref{related}.
We describe BERTeam in Section \ref{methods} and our experiments in Section \ref{experiments}. 
We describe results in Section \ref{results}, and our takeaways in Section \ref{conclusion}.
}




\section{Related Works}\todo{While it's true that having a fixed set of agent policies or no longer training against some previously seen teams can diminish the agents' generalization capabilities, existing self-play-based methods are equipped with methods to handle that. Team-PSRO [2] equips the team generation method with a mix-and-match operator that combines previously identified teams to create new ones (like this work). At the same time, methods like Fictitious Co-play [4] save past self-play checkpoints and still interact against them to prevent forgetting. (YQqx)}

\textbf{Self-Play:}
Self-play is a central concept for training autonomous agents for adversarial games. 
The main idea is to keep track of a set of policies to play against during training \cite{selfplay}.
These policies are usually current or past versions of agents being trained. 
Extending the concept of self-play, Alpha-Star \cite{jaderberg2019human} utilized league play, a technique where a diverse set of agents with different performance levels play in a tournament style game structure.
League play improved the adaptability of trained agents to play against different opponent difficulties in the Starcraft-II real-time strategy video game. 
Similar to the league-play concept, we use coevolution in our proposed technique to train agents in games against a variety of opponent strategies and skill levels.

\textbf{Team Selection in Adversarial Games:}
In \cite{mcaleer2023team}, researchers proposed Team-PSRO (Policy Space Response Oracle), a technique within a framework called TMECor (Team Max-min Equilibrium with Correlation device), for multi-player adversarial games. 
In Team-PSRO, agents improve their policies iteratively through repeated game-play. 
In each iteration, the policy for each agent in a team is selected as a component of a best-response policy for the entire team, calculated by a best response oracle. 
In our technique, the role of this oracle is performed by BERTeam's team selection. 

Another approach for learning to play adversarial team games trains agents incrementally via curriculum learning, then adapts those learned strategies for adversarial settings via self-play \cite{lin2023tizero}. 
Strategies are stochastically modified to introduce diversity and to increase adaptability against unseen opponent strategies.
While this technique adapts previously trained policies for adversarial play, our approach creates a population of policies trained in an adversarial environment, then selects the best team from this population. 
The large number of potential teams to select from allows us to adapt our team selection to newer opponent strategies.

\textbf{Evolutionary Algorithms for Multiagent Games:}
Evolutionary algorithms have been used for decades for multiagent games due to their adaptability and performance in domains like soccer \cite{footballteammemberselection}.
Coevolution in particular \cite{coop_co,coop_co2} has the advantage of evolving independent populations of agents for specialized skills (e.g. defending, passing, shooting).
However, a downside to this is that the correct specialized agents must be chosen for each environment.

To address this, authors in \cite{teamformationassesor} proposed a technique to use an assessor to choose team compositions intelligently. 
The assessor was a model trained to predict the outcome of games between known opponents.  
This allowed them to evolve specialized agents while also being able to choose optimal teams against known opponents. 
However, to select the optimal team against a known opponent team, they must search the space of all possible teams, querying their model each time. 
To select a good team against a partially known opponent team, they repeatedly update the current and opponent teams, continuing until convergence. 
Thus, this causes difficulty in scalability to larger populations, and reduces adaptability in cases where  opponent policies are not known a priori.

In their research, Dixit et al. achieve a similar goal of diversifying individual policies and selecting an optimal team composition from these diverse agents \cite{Dixit22}.
To diversify agents, they used \gls{QD} methods like MAP-Elites, an evolutionary algorithm that ensures the population evolved has sufficiently diverse behavior \cite{mapelite}. 
MAP-Elites works by projecting each policy into a low dimensional behavior space, then considering only fitnesses of agents that behave similarly when updating the population.

Dixit used \gls{QD} on a few independent islands of \gls{RL} agents. 
To select optimal team compositions, their \textit{mainland} algorithm keeps track of a distribution of the proportion of members from each island that compose an optimal team. 
To train this distribution, they repeatedly evaluate agents in team games, then use the outcomes to rank the teams. 
They use the compositions of the best few teams to update the distribution, and the individual fitnesses and \gls{RL} training examples to update the individual policies on each island.

MCAA is designed and evaluated for cooperative tasks like visiting a set of targets with robots that have different navigation capabilities (e.g. drones and rovers). 
Similar to MCAA, our proposed algorithm uses two separate components for evolving agent skills and evaluating agent performance. 
However, we diverge from MCAA as we consider adversarial teams and cases where the agents are uniform and only differentiated by the behavior of their learned policies. 
We compare the main techniques of our proposed algorithm with analogous components of MCAA.

\textbf{Transformers:}
Transformers are a sequence-to-sequence deep neural network architecture designed to create context-dependent embeddings of input tokens in a sequence \cite{trans}. 
They use an encoder-decoder architecture \cite{encdec1,encdec2,encdec3}, taking in an input and a target sequence. 
The output is an embedded sequence corresponding to each element of the target sequence. 
A final layer is often added, converting each element into a probability distribution over a vocabulary to allow sequence generation.
This architecture is widely used in \gls{NLP}, for various tasks including generation, classification, and translation \cite{generation,classification1,translation2,translation1}.

\gls{BERT} \cite{bert} is an update to the original transformer \lrquote{next token prediction} training structure. 
\gls{BERT} is instead trained with \gls{MLM}, inspired by the Cloze task \cite{cloze}. 
This forces the model to predict randomly masked tokens given bidirectional context, which improves robustness \cite{bert, bertrobust}.
We use a similar approach, associating a set of agents to tokens and using an \gls{MLM} training scheme to produce sequences of agents to form strong teams.

\section{Team Selection in Adversarial Games}\textbf{Preliminaries}:
A \gls{MDP} is a framework capturing a broad range of optimization tasks.
An \gls{MDP} is described by a tuple $( S, A, \mathcal T,\mathcal R,\gamma)$,
containing a state space $S$,
an action space $A$,
a transition function $\mathcal T(S\mid S\times A)$, 
a reward function $\mathcal R\colon S\times A\times S\to \mathbb R$, 
and a discount factor $\gamma\in [0,1)$.
An agent is a player in an \gls{MDP} and is described by its policy $\pi(A\mid S)$.
The sequence $(s_0,a_0,r_0),(s_1,a_1,r_1),\dots$ is referred to as a trajectory,
where $a_i$ is sampled from $\pi(a\mid s_{i-1})$, 
$s_i$ is sampled from the transition function $\mathcal T(s\mid s_{i-1},a_{i-1})$, 
and $r_t=\mathcal R(s_t,a_t,s_{t+1})$.
An agent's objective is optimizing the sum of discounted rewards in a trajectory:
$\mathbb E[\sum\limits_{i}\gamma^ir_i]$.

\todo{Adversarial games can only be formulated as a POMDP (from the viewpoint of a single agent) if other agents deemed part of the environment do not change their policies (i.e., not learning or adaptive to the trained agents' actions). This is certainly a strong assumption for the class of opponents that can be considered in these types of work. A better formulation (that supports setting different rewards for different agents) could have been based on Stochastic/Markov Games. (YQqx)
}

Reinforcement Learning (\gls{RL}) is one method of optimizing a policy for an \gls{MDP}. 
While there are various techniques for different classes of \gls{MDP}s, most of them keep track of a policy $\pi_\theta(A\mid S)$ and a value estimator $r_\phi\colon S\to \mathbb R$. 
After sampling an episode, the value parameters $\phi$ are updated towards the observed discounted rewards from each state. 
The policy parameters $\theta$ are updated using the return estimator to optimize the expected discounted rewards from each state.
Deep \gls{RL} utilizes Deep Neural Networks to create policy and value networks $\pi_\theta,r_\theta$.

We are concerned with multiagent $k$-v-$k$ adversarial games. 
Given the policies of all agents playing, this scenario can be formally defined as an instance of an \gls{MDP} for each agent, with a distinct reward function for each agent. 
At each time step, the actions of all other agents are considered in the transition function $\mathcal T$. 
We divide the agents into teams and additionally define an outcome evaluation that considers the trajectories of a game and returns a set of teams that \lrquote{won}.\footnote{A natural extension of this is to have real number outcomes for each team. 
We discuss a training method for this situation in Appendix \ref{nashpendix}}
We assume the \gls{MDP} rewards of an agent correlate with its team's outcome, so agents that get high rewards in a game are likely to be on winning teams. 
We use this framework as opposed dec-MDPs, a formalization used in team reach-avoid games \cite{decmdp, shapley1953stochastic, reachavoid1, reachavoid2} since we would like each agent to have their own reward structure, and for these rewards to be separate from the game outcomes. 
Within our framework, the problem we consider is how to best create a team of agents whose policies are likely to win against a variety of opponents.

\todo{The optimized objective (i.e., maximizing returns against unknown teammates) is not formalized well. The authors should provide a mathematical formulation of their objective. In addition, the authors should try to clarify whether they assume certain assumptions regarding opponents' behavior seen in the evaluation. (YQqx)}

This problem is difficult to solve due to the cooperation required of policies within a team and the variety of potential opponent policies.
There are two general directions to address this problem. 
The first method maintains a fixed number of agents per team corresponding to the team size; each agent's policy is trained iteratively via techniques like self-play \cite{lin2023tizero, mcaleer2023team}. 
However, maintaining a fixed set of agent policies limits the adaptability of each agent as well as the diversity of the team. 
It might be difficult to play successfully against an opponent strategy that may have been \lrquote{forgotten} during training.
The second approach \cite{jaderberg2019human}, which we adopt in this research, is to maintain a larger set of agent policies and select a few agent policies to form a team.
This creates vast diversity in possible teams with the downside of introducing the additional overhead of team selection.
This is a non-trivial problem as the team selection must select policies that cooperate well, while remaining cognizant of the opponent's possible strategies.
At the same time, policies must be continuously improved via self-play learning to remain competitive against newer opponent strategies.

Our proposed technique is to use coevolution to train a large set or population of agent policies, and use a transformer-based sequence generation technique to select optimal teams from the population.
These two techniques are illustrated in Figure \ref{berteamcoevolution} and described in more detail in the following sections.

\subsection{Transformer based Sequence Completion for Team Selection}
{\em Why Transformers? }
Transformers are generative models that can be used to query \lrquote{next token} conditional distributions. 
A size $k$ team can be generated in $k$ queries, giving us efficient use of the model.
The form of the output also allows us make specialized queries for cases where valid teams may have specific constraints.\footnote{An example of this would be a team that must be partitioned by agent types. In this case, we would use the output to sample each member from a set of valid choices.
} 
Transformers are also able to condition their output on input embeddings, allowing us to form teams dependent on observations (of any form) of the opponent.
We may also pass in an empty input sequence for unconditional team selection.
We expect that through their initial embeddings, transformers will encode similarities between agents and allow the model to infer missing data. 
This is supported by the behavior of word  embeddings in the domain of NLP \cite{word2vec}. 

One drawback with transformers is their $\Theta(k^2)$ complexity for a sequence of size $k$. 
However, sequence lengths of up to 512 are easily calculated \cite{bert}, indicating that our algorithm can scale up to this team size.\footnote{This would come with an increase in training data required for sensible output.} 
Even for team sizes beyond this limit, there exist workarounds like sliding-window attention \cite{longformer}.



\begin{figure}[htb!]
    \centering
    \includegraphics[width=\linewidth]{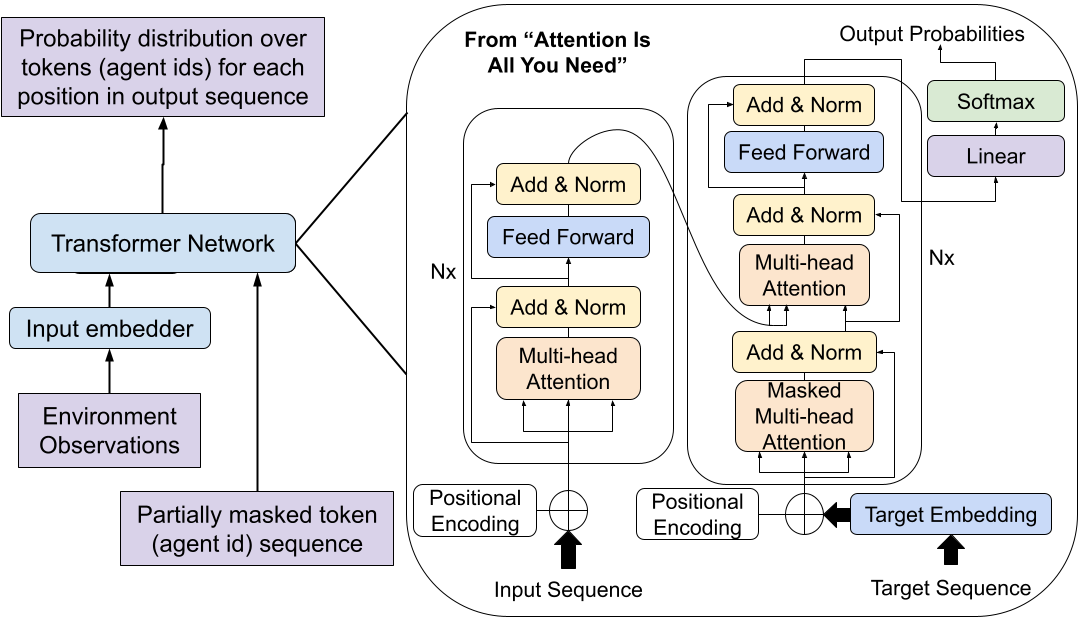}
    \caption{BERTeam's core, a transformer network}
    \Description{Transformer network architecture, taken from the original paper, along with its use in BERTeam.}
    \label{fig:trans_architect}
\end{figure}

\subsubsection{Model Architecture}

The core of BERTeam is a transformer model whose tokens represent each possible agent in the population, along with a [MASK] token.
A separate input embedding model transforms observations of any form into a short sequence of vectors on which to condition BERTeam's output.\footnote{
While this is a capability of BERTeam, we choose to analyze BERTeam as an unconditional team generator. 
We plan to explore input embeddings in future work.
} 
Since the architecture of this input embedding depends on the form of the environment observations, this model must be tailored to each use case. 
A query of the model will take as input a partially masked team and any environment observations. 
The output is a predicted distribution over agents for each element of the sequence, as illustrated in Figure \ref{fig:trans_architect}.
During evaluation, the model will be used as a generative pre-trained model (as in \cite{trans}) to sample a team.

During training, we use BERTeam to sample opponent teams as well.
In this case we may include the team index as an input to condition BERTeam on, or simply use a different instance of BERTeam for each team.
However, there is often symmetry between teams, which makes this unnecessary.

\subsubsection{Training Procedure}

The BERTeam model uses \gls{MLM} training, which requires a dataset of \lrquote{correct sequences}. The model is trained to complete masked sequences to match this distribution.

To generate this dataset, we consider the outcomes of games played between various teams. Since our goal is for the model to predict good teams, we fill the dataset with only teams that win games, along with the observations of their players. We also include examples of winning teams without any observations to train the model to generate unconditioned output.

Since it is usually infeasible to sample all possible team pairs uniformly, we utilize BERTeam to sample teams and their opponents.
The motivation of this is to fill the dataset with teams that win against well coordinated opponents. 
Additionally, under certain assumptions and dataset weights, training BERTeam on this dataset will provably converge to a Nash Equilibrium. 
We describe this in Appendix \ref{nashpendix}, but do not implement this in our experiment, as it may cause poorly behaved learning.

Thus, as in Figure \ref{berteamcoevolution}, we train BERTeam alongside generating its dataset, and utilize the partially trained model to sample games. Our dataset takes the form of a replay buffer, so outdated examples are gradually replaced.
One concern is that filling the dataset with teams generated from BERTeam may result in stagnation, as the dataset will match trends in the distribution. 
To mitigate this, we use inverse probability weighting \cite{likelihood}, weighting rare samples more. 
We discuss this in Appendix \ref{weightpendix}.

\subsubsection{Training along with Coevolution}
\label{berteam_coevolution_training_elo_stuff}
\begin{figure}[ht!]
    \centering
\includegraphics[width=.88\linewidth]{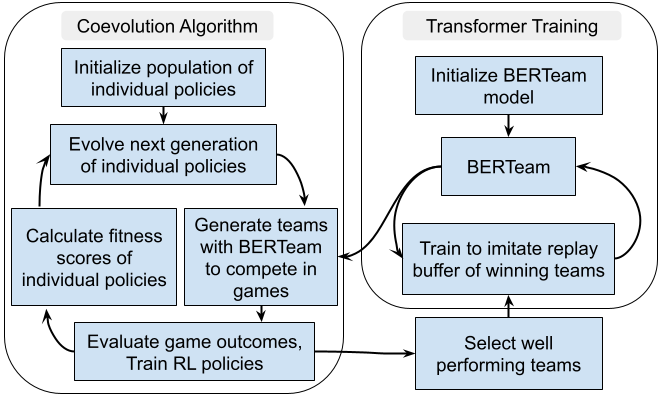}
    \caption{Training of BERTeam alongside coevolutionary RL}
    \Description{Diagram of BERTeam algorithm for the use case of training alongside agents that update their policy.}
    \label{berteamcoevolution}
\end{figure}
The BERTeam model is able to be trained alongside coevolution, as its past knowledge of the agent policies can be utilized and updated with recent information.
The outcomes of games sampled in coevolution can be used in the dataset of BERTeam, and the BERTeam partially trained model can be used to sample better teams, a cycle displayed in Figure \ref{berteamcoevolution}. 
To use a coevolution algorithm (detailed in Section \ref{coev_section}) we must define an individual agent fitness function using BERTeam with sampled games.
This is tricky, as we have only assumed the existence of a comparative \textit{team} outcome. 
We solve this by utilizing Elo, a method of assigning each player a value from the results of pairwise 1v1 games \cite{elo}.
Given a \textit{team selector} (i.e. BERTeam) that can sample from the set of all possible teams containing an agent (the \textit{captain}), we define the fitness of each agent as the expected Elo of a team chosen with that agent as captain. 
To update these values, we sample teams to play training games, choosing our captains by adding noise to BERTeam's initial distribution. 
We update the fitnesses of each team captain with a standard Elo update (Appendix \ref{elopendix}). 
Since it is confusing to distinguish these individual fitnesses from team Elos, we will refer to them as fitness values from now on.
We do not update fitnesses of non-captain members since their evaluation may be skewed by teammates who were chosen unconditional to them being on the team.
The captain does not have this issue since the team is selected knowing the captain is a member.


\subsection{Coevolution For Training Agent Policies}
\label{coev_section}
\begin{algorithm}[htb!]
\SetAlgoLined
\SetKwInOut{Input}{input}\SetKwInOut{Output}{output}
\SetKwProg{myproc}{Procedure}{}{}
\Input{$Pop$: agent population\\
$k$: size of each team\\
}
\Output{$Pop$: updated agent population via coevolution}
\BlankLine
\SetKwFunction{proc}{train-pop-coevolution}{}
\myproc{\proc{$Pop, k$}}{
$f_{i} \leftarrow 1000\;\;\forall i\in Pop$
\label{alg_line:fitness_init}\\
\For {$1 ... n_{epochs}$}{
    $\mathcal T_i\leftarrow \emptyset\;\;\forall i\in Pop$
        \label{alg_line:trajectory_init}\\
    \For {$1,\dots,n_{games}$\label{alg_line:game_sample}}{
        $(T, T') \leftarrow$  sample $k$ agents per team from $Pop$ using BERTeam
        \label{alg_line:team_sample}\\
        
        $cap, cap' \leftarrow$ select captains for teams $T,T'$
        \label{alg_line:captian_select}\\

        $g\gets$ game played between teams $T,T'$
        \label{alg_line:game_playing}\\

        $O\gets$ outcomes for teams $T, T'$ from game $g$
        \label{alg_line:outcome_collect}\\
        
        $\mathcal T_i\gets \mathcal T_i\cup$ trajectories from $g$ for agent $i,$ $\forall i \in \{T\cup T'\}$
        \label{alg_line:trajectory_collect}\\

        $f_{cap}, f_{cap'} \leftarrow$ update fitness($cap, cap',O$)
        \label{alg_line:fitness_update}\\
    }
    Update $\pi_i$ with RL using experience in $\mathcal T_i \;\;\forall i\in Pop$
    \label{alg_line:rl_step}\\
    
    $\mathbf{P}_{clone} \leftarrow$ Clone $n_{rem}$ agents from $Pop$ selected using Eqn. \ref{eqn:agent_selection_coev}
    \label{alg_line:clone_select}\\
    
    $\mathbf{P}_{rem} \leftarrow$ Select $n_{rem}$ agents from $Pop$ using Eqn. \ref{eqn:agent_removal_coev}
    \label{alg_line:remove_select}\\
    
    $Pop \leftarrow \{Pop \setminus \mathbf{P}_{rem}\} \cup \mathbf{P}_{clone}$\label{alg_line:pop_update}
    
    }
    return $Pop$
}
\caption{Algorithm for training a population of agents via coevolution self-play}
\label{algo:train_pop_coev}
\end{algorithm}


The main idea in coevolution is that instead of optimizing a single team, 
a population of agents learns strategies playing in games between sampled teams. 
We choose to use Coevolutionary Deep RL since it allows agents to be trained against a variety of opponent policies, addressing performance against an unseen opponent.

Thus, we use Algorithm \ref{algo:train_pop_coev}, heavily inspired by \cite{coevdeeprl}, to produce a population of trained agents.
The input is an initialized population of agents, as well as parameters like team size. 
In each epoch, we sample $n_{games}$ games, which each consist of selecting teams and captains using a team selector (lines \ref{alg_line:team_sample}, \ref{alg_line:captian_select}), 
playing the game (line \ref{alg_line:game_playing}), 
and collecting trajectories and outcomes (lines \ref{alg_line:outcome_collect}, \ref{alg_line:trajectory_collect}). 
The outcomes are sent to the team selector for training and also used to update individual agent fitnesses in line \ref{alg_line:fitness_update} (see Figure \ref{fig:trans_architect}).
At the end of each epoch, the trajectories collected update each agent policy in place in line \ref{alg_line:rl_step}, and the population is updated in lines \ref{alg_line:clone_select}-\ref{alg_line:pop_update}. 

The parts that deviate from the coevolutionary \gls{RL} algorithm in \cite{coevdeeprl} are the fitness updates (lines \ref{alg_line:captian_select} and \ref{alg_line:fitness_update}) and the generation update (lines \ref{alg_line:clone_select}-\ref{alg_line:pop_update}). 
The deviation in fitness updates is a result of only assuming the existence of a \textit{team} evaluation function, and is discussed in Section \ref{berteam_coevolution_training_elo_stuff}.
The deviation in population updates is due to considerations with training alongside BERTeam. 
BERTeam assumes similarities in behavior of the agent assigned to each token, and if we replace or rearrange every member of the population, the training of the BERTeam model would be rendered useless.
By controlling replacement rate, we ensure most of the information learned by BERTeam retains relevance. 
To do this while best imitating \cite{coevdeeprl}, we stochastically choose $n_{rem}$ agents to replace and $n_{rem}$ to clone\footnote{An agent might be selected for both replacement and cloning, resulting in no change.} using the following equations:

\begin{eqnarray}
\label{eqn:agent_selection_coev}
    \mathbb P\{\text{clone agent }i\}=\frac{\exp(f_i)}{\sum\limits_{j\in Pop}\exp(f_j)} \\
\label{eqn:agent_removal_coev}
    \mathbb P\{\text{remove agent }i\}=\frac{\exp(-f_i)}{\sum\limits_{j\in Pop}\exp(-f_j)} 
\end{eqnarray}

Additionally, we may tune hyperparameters such as RL learning rate so that each individual policy update is minor and keeps an updated agent's behavior similar to its predecessor.\footnote{This additionally justifies 
a shortcut in the proposed algorithm, which uses the same sampled games to update policies through RL and to update fitness values. 
An alternate approach would be to sample $n_{games}$ twice per epoch, once to update policies, and once to update fitnesses, which is unnecessary if each update is minor.}

\section{Experiments}
To better analyze the effectiveness of this algorithm, we consider 2v2 team games. 
This small team size allows us to more easily analyze the total distribution learned by BERTeam.

\subsection{Aquaticus}

\begin{figure}[htbp]
\centering
\subfigure[MOOS-IvP Environment]{
\includegraphics[height=50 pt]{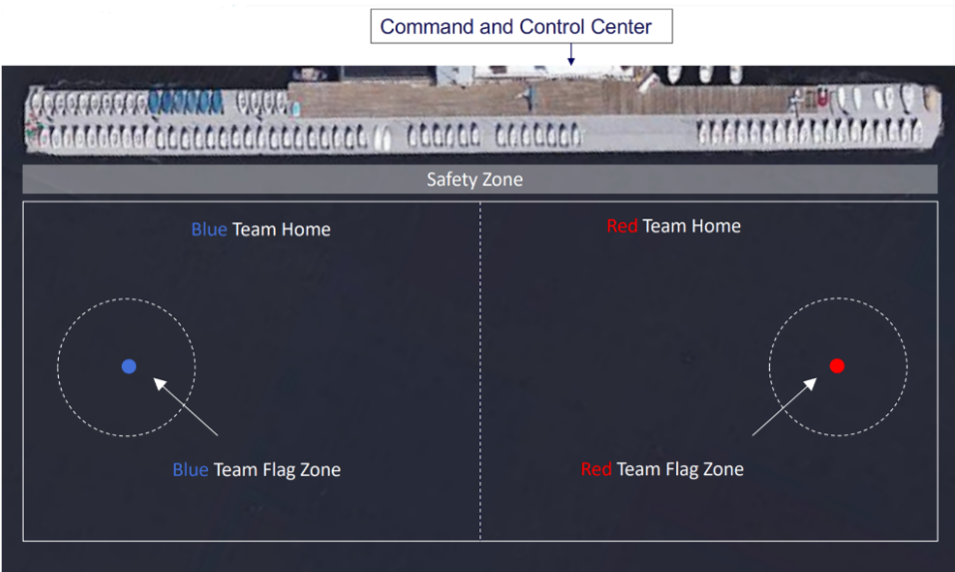}
}\subfigure[Pyquaticus environment]{
\includegraphics[height=50 pt]{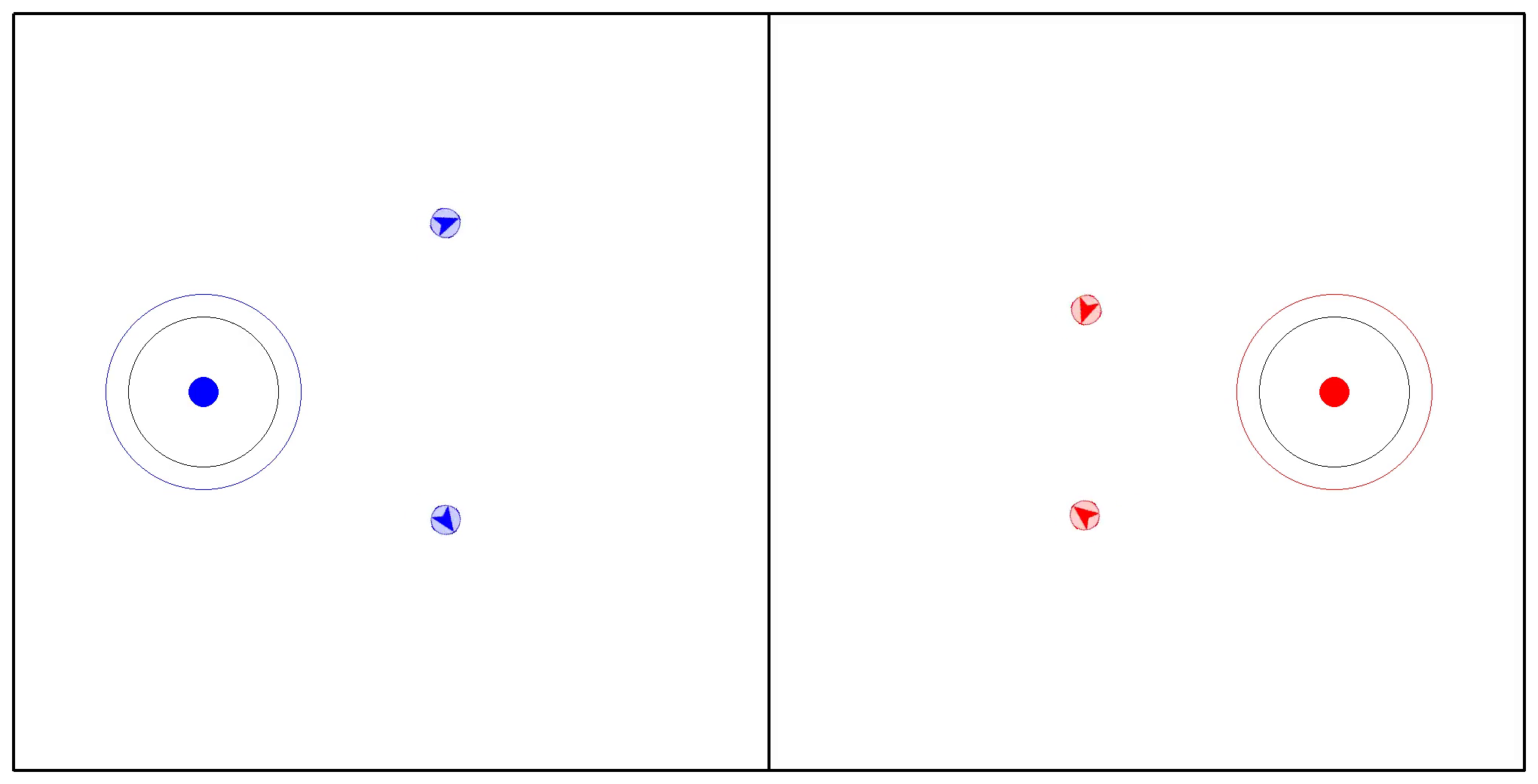}
}
\caption{Aquaticus game, and its simulated version}
\label{quatici}
\Description{Physical Aquaticus environment and simulated Pyquaticus environment.}
\end{figure}

Aquaticus is a Capture-the-Flag competition played with teams of autonomous boats \cite{aquaticus}.
We are interested in this task because it is an example of a multiagent adversarial team game.
Each agent has low-level motor control, and thus any complex behaviors must be learned through a method like \gls{RL}.
Additionally, there are various strategies (e.g. offensive/defensive) that agents may adopt that perform well in competition.
Finally, we believe that optimal team composition in this game is non-trivial, and expect that a good team is composed of a balanced set of strategies. 

\subsubsection{Pyquaticus}
Due to the difficulty of testing on real robotic platforms, we test and evaluate our methods on Pyquaticus, a simulated version of Aquaticus \cite{pyquaticus}. 
The platform is implemented as a PettingZoo environment \cite{pettingzoo}, the standard multiagent extension of OpenAI Gymnasium \cite{openaigym}.
In experiments, we use the \gls{MDP} structure implemented in Pyquaticus. 
We terminate an episode when a team captures a flag, or after seven in-game minutes. 

\subsection{Team Selection with Fixed Policy Agents}
\label{BERTeam_exp}

To test the effectiveness of BERTeam independent of coevolution, we use fixed policy agents predefined in Pyquaticus: a random agent, three defending agents, and three attacking agents.
The attacking and defending agents each contain an easy, medium, and hard policy. 
The easy policies move in a fixed path, and the medium policies utilize potential field controllers to either attack the opponent flag while avoiding opponents, or capture opponents by following them. 
The hard policies are the medium policies with faster movement speed.
We follow the training algorithm in Figure \ref{berteamcoevolution} without the coevolution update.
We conduct an experiment with the 7 agents, and detail experiment parameters in Appendix \ref{parampendix}.

Throughout training, we record the occurrence probability of all possible teams using BERTeam. 
We do this exhaustively, by considering all possible sequences drawn from the set of agents with replacement. 
We expect that as BERTeam trains, this will gradually approach a distribution that favors well performing teams.

\subsubsection{Elo Calculation}
\label{trueelo}
Since we have a fixed set of policies, we can compute true Elo values for all 28 unordered 2 agent teams (see Appendix \ref{number_of_possible_teams}), obtained from an exhaustive tournament of all possible team pairings. 
For these results, we perform 10 experiments for each choice of two teams. 
We use the scaling of standard chess Elo \cite{elo}, and shift all Elos so that the mean is 1000.

\subsection{Team Selection with Coevolution of Agents}
\label{coevolution_exp}

We train BERTeam alongside Algorithm \ref{algo:train_pop_coev}, as illustrated in Figure \ref{berteamcoevolution}.\footnote{
Our implementation, along with experiments, is available at \cite{BERTeam, coevolver}
}  
For each individual agent, we use \gls{PPO}, an on-policy \gls{RL} algorithm \cite{ppo}.
We detail experiment parameters in Appendix \ref{parampendix}.
To find the Elos of each possible team, we utilize the Elos calculated for the fixed policy teams as a baseline. 
For each of the 1275 possible teams,\footnote{We use a population of 50 agents and a team size of 2, see Appendix \ref{number_of_possible_teams}.} we test against all possible teams of fixed policy agents. 
We then use the results of these games to calculate the true Elos of our teams of trained agents. 
We do not update the Elos of our fixed policy agents during this calculation.

For policy optimization, we use stable\_baselines3 \cite{stable-baselines3}, a standard RL library. 
Since this library is single-agent, we create unstable\_baselines3,\footnote{
Code available at \cite{unstable_baselines3}
} a wrapper allowing a set of independent learning algorithms to train in a PettingZoo multiagent environment.

\subsubsection{Aggression Metric}
\label{aggression}
Without a method to classify behavior of individual learned policies, the output of BERTeam is difficult to interpret.
Thus, we create an aggression metric to estimate the behavior of each agent in a game. 
Specifically, this metric for agent $a$ is $\frac{1+2(\# a \text{ captures flag})+1.5(\# a\text{ grabs flag})+(\# a\text{ is tagged})}{1+(\# a\text{ tags opponent})}$, where $(\# a\text{ event})$ denotes the number of times that event happened to agent $a$ in the game.
We evaluate the aggressiveness of each trained agent by considering a game where one team is composed of only that agent.
We evaluate the average aggression metric against all possible teams of fixed policy agents. 
We expect aggressive agents to have a metric larger than 1, and defensive agents to have a metric less than 1.\footnote{We tuned our metric to distinguish our fixed policy agents.}


\subsection{Comparison with MCAA}
\label{comparison_exp}
We notice that the \gls{MCAA} mainland team selection algorithm is playing an analogous role to BERTeam team selection, and that the MAP-Elites policy optimization on each island is analogous to our coevolutionary \gls{RL} algorithm.
Since both our algorithm and \gls{MCAA} distinguish team selection and individual policy optimization as separate algorithms, we may hybridize the methods and compare the results of four algorithms.
The algorithms are distinguished by choosing MAP-Elites or Coevolutionary Deep RL for policy optimization, and BERTeam or \gls{MCAA} for team selection.
The hybrid trials will allow us to individually evaluate BERTeam as a team selection method, independent of the policy optimization.

We must make some changes to be able to implement \gls{MCAA} and MAP-Elites in an adversarial scenario.
First, a step of the \gls{MCAA} algorithm ranks a set of generated teams with a team fitness function. 
To approximate this in an adversarial environment, we play a set of games each epoch, and consider the teams that won as \lrquote{top ranked}, and include them in the \gls{MCAA} training data.
Similarly, \gls{MCAA} assumes an evaluation function for fitness of individual agents.
We approximate this by considering the teams each individual has been included in. 
We take a moving average of the comparative team evaluations, and assign this average to the selected agent.

For MAP-Elites, we adapt our aggression metric as a behavior projection. 
While this could be made more complex, we believe this is sufficient, as our work is focused on the team selection aspect.
Additionally, we adapt the MAP-Elites algorithm to remove $n_{rem}$ poorly performing agents from the population:

\begin{itemize}
    \item Project each policy into a low dimensional behavior space. Let the behavior vector of agent $a$ be $B_a$.
    \item Protect \lrquote{unique} agents from deletion. Given a neighborhood size $\lambda$, we consider agent $a$ unique if $B_a$ is $\lambda$-far from any other $B_{a'}$. 
    We may increase $\lambda$ if too many agents are unique.
    \item Obtain fitness scores $f_a$ for each agent, and fitness predictions $f'_a$ using each agents neighborhood average.
    \item Delete $n_{rem}$ agents stochastically using Equation \ref{eqn:agent_removal_coev} on relative fitness $f_a-f'_a$. 
\end{itemize}
We do this to keep the spirit of MAP-Elites while allowing it to maintain a fixed population size, which is necessary for the hybrid trial with BERTeam.
The original MAP-Elites algorithm can be recovered by removing stochasticity and repeatedly running our version with fixed $\lambda$ until every agent is unique. 

Another consideration is that in the \gls{MCAA} paper, islands were distinguished by having different proportions of various types of robots (i.e. drones and rovers). 
In our case, we cannot do this as all agents are homogeneous. 
To imitate this variation, we implement a different \gls{RL} algorithm on each island, varying the algorithm type as well as the network size used. 
We describe these algorithms, along with other experiment parameters in Appendix \ref{parampendix}.

Once all four algorithms have been trained, we fix the learned agent policies and team selection policies.
We find the expected Elo ($\mathbb E[\text{Elo}]$) of an algorithm with the expected Elo of a team sampled from the algorithm's team selector.
We evaluate the relative performance of two algorithms by sampling teams and evaluating the games played. 
We sample 10000 games for each of the 6 algorithm comparisons.
We additionally ground our Elo estimates by sampling 1000 games against our fixed policy teams.
When doing Elo calculations, we do not update the fixed policy teams.
We sample since for a population size of $n$, there are $\Theta(n^2)$ possible teams, resulting in an infeasible $\Theta(n^4)$ possible pairings between algorithms. 

\section{Results}
In MCTF, reordering the members of a team has no effect on the team's performance. 
However, BERTeam is a sequence generation model, so it does distinguish order. 
To make results more readable, we assume any reordering of a given team is equivalent to the original team, and calculate distributions and Elos accordingly. 
A caveat to this is that teams with two distinct agents are counted twice in a total distribution, while teams with two copies of one agent are counted once.
To compensate for this during analysis, we double the distribution value of the second type of team, and normalize.
This is relevant mainly in Figure \ref{total_top_10}(a) and Table \ref{rankings2v2}, where we inspect the total distribution of a small number of teams. 
For the analysis of cases with many agents, this effect is negligible.

\subsection{Team Selection with Fixed Policy Agents}


\begin{figure}[ht!]
    \centering
\subfigure[Total distribution]{
\includegraphics[width=.5\linewidth]{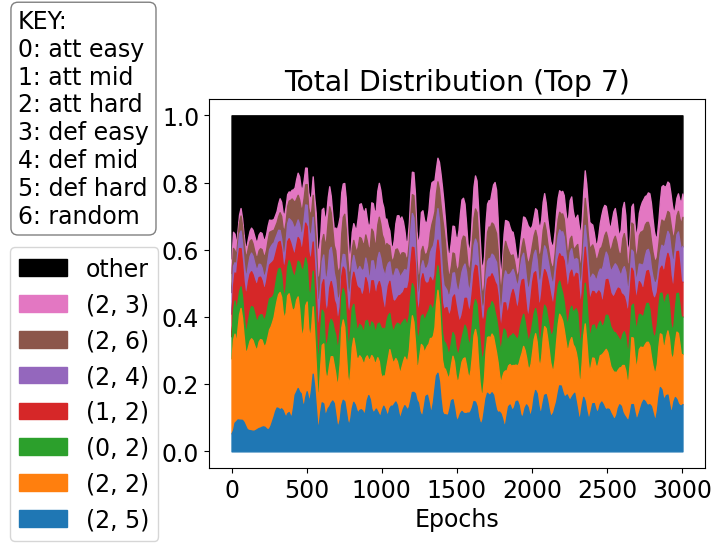}
}\subfigure[Captain distribution]{
\includegraphics[width=.5\linewidth]{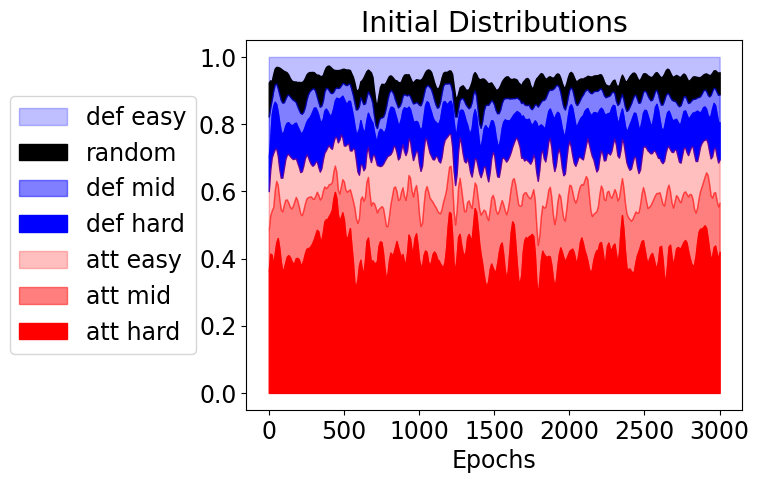}
}
\caption{BERTeam distributions throughout training, sorted by probability (largest on bottom)}
\Description{BERTeam distributions throughout training, sorted by probability}
\label{total_top_10}
\end{figure}

Through training, we inspect the total distribution learned by BERTeam.
From Figure \ref{total_top_10}(a), we notice that our training algorithm learns a non-uniform distribution, as the top seven out of 21 possible team compositions account for about 75\% of the total distribution. 
BERTeam seems to immediately favor teams containing the hard attacking agent, as the teams containing agent 2 are favored in the first few epochs.
Initially, BERTeam's distribution favored $(2,2)$ (orange), the team composed of only hard attacking agents.
However, around epoch 500, the distribution shifted and team $(2,2)$ sharply decreased in occurrence probability, in favor of team $(2,5)$ (blue), the strong balanced team.
After this, there were no major changes in BERTeam's output distribution. 


\begin{table}[ht!]
    \centering
    \begin{tabular}{|c|c|c|c|c|}
    \hline
    \multirow{2}{*}{\textbf{Team}} & \textbf{True} & \textbf{True} & \textbf{Predicted} & \textbf{BERTeam} \\
     & \textbf{Rank} & \textbf{Elo} & \textbf{Rank} & \textbf{Occurrence} 
    \\\hline
    (2, 5) & 1 & 1388 & 1 & 0.14
    \\\hline
    (2, 2) & 2 & 1337 & 2 & 0.13
    \\\hline
    (2, 3) & 3 & 1135 & 7 & 0.06
    \\\hline
    (1, 2) & 4 & 1112 & 4 & 0.10
    \\\hline
    (0, 2) & 5 & 1097 & 3 & 0.10
    \\\hline
    (2, 4) & 6 & 1087 & 5 & 0.10
    \\\hline
    (2, 6) & 7 & 1035 & 6 & 0.07
    \\\hline
    (0, 5) & 8 & 975 & 13 & 0.03
    \\\hline
    \end{tabular}
    \caption{Comparison of true ranks and predicted ranks}
    \label{rankings2v2}
\end{table}

To inspect the performance of BERTeam's favored team compositions, we consider the true Elos of each team. 
In Table \ref{rankings2v2}, we list the eight best performing teams and their true Elos alongside the rankings and occurrence probabilities from BERTeam.
We find that the top two choices made by BERTeam are correct, being the balanced $(2,5)$, and the aggressive $(2,2)$ respectively. 
Their occurrence probabilities $(14\%, 13\%)$ are also reasonably larger than the third rank at $10\%$. 
BERTeam does correctly select the next five, though the order is shuffled. 
With the exception of team $(2,3)$, they all are reasonably close to their correct positions. 

In Figure \ref{total_top_10}(b), we we consider the distribution of team captains chosen by BERTeam. 
This is the expected output distribution of an initial agent given a completely masked sequence. 
We find that agent 2 is strongly favored throughout training, indicating that it is a likely choice in a top performing team. 
This can be related to \gls{NLP}, with an analogous problem of generating the first word of an unknown sentence. 
Just as articles and prepositions (e.g. \lrquote{The}) are strong choices for this task, BERTeam believes agent 2 is a strong choice to lead a team.
This is supported by the top seven teams in Table \ref{rankings2v2} being all teams containing agent 2.
Thus, just as in \gls{NLP}, BERTeam is able to accurately determine agents likely to be in well-performing teams, and choose them as team captains.

These results indicate BERTeam is able to learn a non-trivial team composition, as it predicted top teams with reasonable accuracy. 

\subsection{Team Selection with Coevolution of Agents}

\begin{figure}[htbp!]
    \centering
\subfigure[Clustering based on aggression]{
\includegraphics[width=\halffiguresize]{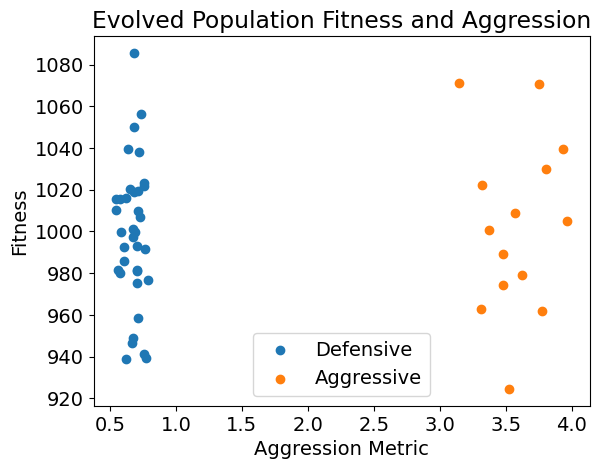}
}\subfigure[BERTeam team composition (ordered)]{
\includegraphics[width=\halffiguresize]{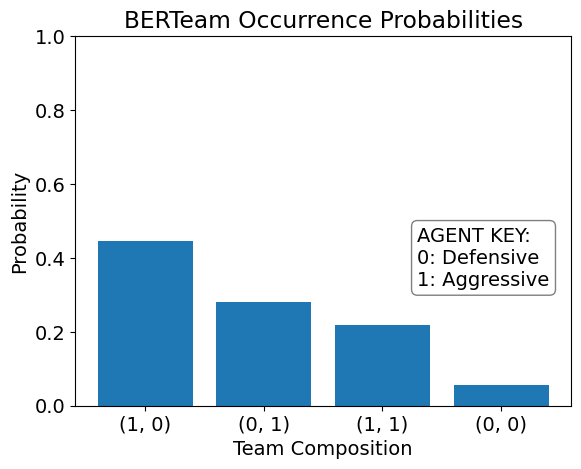}
}
\caption{BERTeam learned distribution on trained agents}
\Description{BERTeam learned distribution on trained agents}
\label{berteam_coev_res}
\end{figure}

After training, we evaluate the evolved agents based on the aggression metric described in Section \ref{aggression}. 
We observe from Figure \ref{berteam_coev_res}(a) that the metric clusters the evolved agents into two clear groups. 
From our population of 50 agents, we classify 36 as {defensive} and 14 as {aggressive}. 
The similar distribution of agent fitness across each population indicates that this diversity is not correlated with agent performance. 
It is instead likely due to specialization for different subtasks.
This indicates that our training scheme supports diversity in agent behaviors during coevolution, even when the \gls{MDP} reward structure for each agent is identical.

We use the grouping of agents by aggression to partition the team distribution learned by BERTeam.
We notice from Figure \ref{berteam_coev_res}(b) that the total distribution of BERTeam heavily favors a balanced team, composed of a defensive and an aggressive member. 
This pairing accounts for about 75\% of the total distribution.
The second most common composition is two aggressive members, accounting for about 20\% of the total distribution. 

The distribution learned by BERTeam aligns with our observation in the fixed policy experiment, where a balanced team performs the best (Table \ref{rankings2v2}). 
This result also implies that BERTeam is learning a non-trivial team composition, since a solution such as \lrquote{always choose the best agent} favors a team composed of one agent type.

\subsubsection{Team Elos}

\begin{figure}[htbp]
    \centering
\includegraphics[width=.69\linewidth]{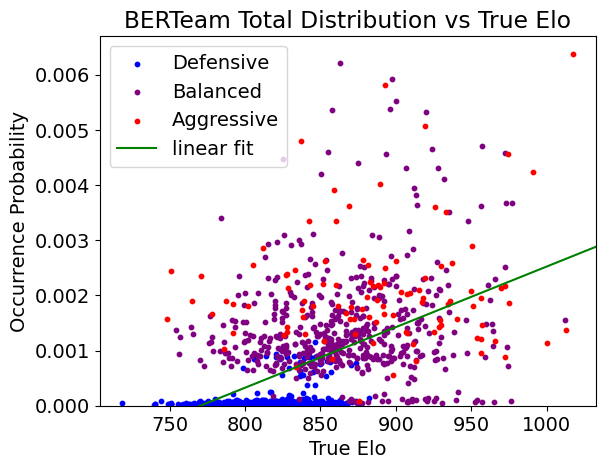}
\caption{BERTeam total distribution and Elos}
\Description{BERTeam total distribution and Elos, partitioned by team composition}
\label{2v2_total_dist}
\end{figure}

We calculate the true Elos for all 1275 teams, using the fixed policy agent teams as a baseline. 
We plot these in Figure \ref{2v2_total_dist}, along with BERTeam's occurrence probability.
We partition all teams into \lrquote{Defensive}, \lrquote{Balanced},  and \lrquote{Aggressive} based on whether they have zero, one, or two aggressive agents.
We also conduct a linear regression on all 1275 teams and plot the line.

We find that there is a correlation with the true Elo of a team and BERTeam's probability of outputting that team.
Our linear regression had a correlation coefficient $R^2\approx .25$, implying that about 25\% of the variance in BERTeam's output is explained by the performance of the team. 
This indicates the distribution learned by BERTeam, while noisy, favors teams that perform well. 

We find that the true Elo of the best performing team is around 1017, indicating a performance slightly better than an average fixed policy team.  
This specific team (composed of two distinct aggressive agents) is also the most probable output of BERTeam. 

Thus, we find that BERTeam, trained alongside coevolution, was able to produce and recognize a team that performed competitively against previously unseen opponents.
In fact, from the rankings in Table \ref{rankings2v2}, we see that the top choice from BERTeam outperforms any team that does not contain agent 2 (the hard offensive agent). 

While the performance of the teams learned from self-play are lower than the top fixed policy agents, this may be a result from difficulties in the environment, such as a lack of correlation between game outcomes and \gls{MDP} rewards. 
This could also potentially be improved by hyperparameter tuning in either the base RL algorithm, the coevolution algorithm, or in BERTeam. 

Overall, the reasonable performance of top coevolved teams, as well as the positive correlation in Figure \ref{2v2_total_dist}, indicate that BERTeam trained alongside coevolution is successful at optimizing policies for a multiagent adversarial team game against unknown opponents. 


\subsubsection{Agent Embeddings}
Recall that the first step of a transformer is to assign each token a vector embedding.
We directly inspect the agent embeddings learned by BERTeam. 
For a subset of the total population, we consider the average cosine similarity of each pair chosen from the subset. 
We use this as an estimate of how similar BERTeam believes agents in that subset are.

For our subset choices, we divide the total population into aggressive and defensive agents, as in Figure \ref{berteam_coev_res}(a). 
We further divide each subset into \lrquote{strong} and \lrquote{weak} based on whether their Elo is above the population average. 
We expect that BERTeam's embeddings of each class of agents have more similarity than a subset of the same size chosen uniformly at random.
We calculate the cosine similarity of a uniform random subset in Appendix \ref{cos_sim_proof}.

\begin{table}[ht!]
    \centering
    \begin{tabular}{|c|c|c|}
    \hline
    \textbf{Subset} & \textbf{Avg Cosine Similarity} & \textbf{Size} 
    \\\hline
    \hline
    Uniform Random & -0.00306 & Any \\\hline\hline
    
    Aggressive & 0.0153 & 14 \\\hline
    Defensive & -0.00449 & 36 \\\hline\hline
    
    Strong Aggressive & 0.01984 & 8 \\\hline
    Weak Aggressive & 0.03846 & 6 \\\hline
    Strong Defensive & 0.00741 & 17 \\\hline
    Weak Defensive & -0.00842 & 19 \\\hline
    \end{tabular}
    \caption{Average cosine similarity of BERTeam's learned initial embeddings across various population subsets}
    \label{cossim}
\end{table}

From the results in Table \ref{cossim}, we can see that the majority of the subsets we chose have a stronger similarity than a random subset of the same size.
The only subsets that do not support this are the \lrquote{Defensive} and \lrquote{Weak Defensive} subsets, which are slightly lower. 

This suggests that the initial vector embeddings learned by BERTeam are not simply uniquely distinguishing each agent. 
The agents that perform similarly are viewed as similar by BERTeam. 
This indicates known properties of token embeddings in the domain of \gls{NLP} (such as word vectors learned in \cite{word2vec}) apply in our case as well. 
The vectors learned by BERTeam encode aspects of each agent's behavior, and similarities in agents can be inferred through similarities in their initial embeddings.
This suggests BERTeam can account for incomplete training data by learning which agents have similar behavior, and thus may be interchangeable in a team.


\subsection{Comparison with MCAA and MAP-Elites}

We directly compare the trained policies using our algorithm, \gls{MCAA}, and the hybrid methods. 
We produce teams using the specified team selection method, and use the outcomes of these games to estimate their expected Elos ($\mathbb E[\text{Elo}]$) in Table \ref{comparison_elos}.

\begin{table}[ht!]
    \centering
    \begin{tabular}{|c|c|c|c|c|}
    \hline
    \textbf{Policy} & \textbf{Team} & \multirow{2}{*}{$\mathbb E[$\textbf{Elo}$]$} & \multicolumn{2}{c|}{\textbf{Avg. update time of}} \\\cline{4-5}
    \textbf{Optimizer} & \textbf{Selection} &  & \textbf{Agents} & \textbf{Team Dist.} 
    \\\hline
    Coevolution & BERTeam & 919 & 13 s/epoch & 46 s/update \\\hline
    Coevolution & \gls{MCAA} & 817 & 13 s/epoch & $\approx 0$ s/update  \\\hline
    MAP-Elites & BERTeam & 883 & 36 s/epoch & 45 s/update \\\hline
    MAP-Elites & \gls{MCAA} & 809 & 35 s/epoch & $\approx 0$ s/update  \\\hline
    \end{tabular}
    \caption{Relative performance of MCAA, our algorithm, and hybrid algorithms}
    \label{comparison_elos}
\end{table}
We find that trials that used BERTeam as a team selection method outperformed trials that used the \gls{MCAA} mainland algorithm. 
One possible explanation for this is the lack of specificity in the \gls{MCAA} mainland algorithm. 
While BERTeam learns the distribution on an individual agent level, \gls{MCAA} chooses the proportion of each island included in a team. 
This method seems to be most effective when the islands are distinct (i.e. in the original \gls{MCAA} paper, islands had different proportions of robot types).
However, in our case, it seems varying the \gls{RL} algorithm did not have a similar effect. 
Another possible explanation is while BERTeam may learn an arbitrarily specific total distribution, \gls{MCAA} can only learn a total distribution that is independent for each position. 
This restricts \gls{MCAA} from learning distributions that do not factor in this way.

Additionally, this supports previous results. When taking the weighted average of Elos in Figure \ref{2v2_total_dist}, we find BERTeam's expected Elo is about 930. 
This is close to the result of 919, and the minor difference can be explained by the fact we trained for less epochs.

\todo{more in depth complexity anal (\autoref{comparison_elos}), or maybe in appendix (WRQK)}

As for runtime complexity, we separately inspect the clock time of team training and of agent policy updates. 
The difference in using BERTeam or \gls{MCAA} to generate teams for policy updates was negligable. The runtime was dominated by conducting the sample games and conducting the \gls{RL} updates. 
For team training, we find the update of \gls{MCAA} took almost no time. In contrast, training the BERTeam model took about 46 seconds on a batch size of 512. 
This significant cost is justified by its stronger performance, and that it can be trained independently without interfering with the flow of the rest of the algorithm. 
Finally, we find our implementation of MAP-Elites is costly, as we sample separate games to perform behavior projection. 
We could mitigate this by implementing MAP-Elites more similar to the original implementation. 

\section{Conclusion}
In this paper, we propose BERTeam, an algorithm and training procedure that is able to learn team composition in multiagent adversarial team games. 
BERTeam is effective both in choosing teams of fixed policy agents and when being trained along with the policies of individual agents. 
It can also take in input, allowing it to generate teams conditional on observations of opponent behavior.

We evaluate this algorithm on Pyquaticus, a simulated capture-the-flag game played by boat robots.
We test BERTeam both with fixed policy agents and training alongside a coevolutionary deep \gls{RL} algorithm. 
We find that in both cases, BERTeam effectively learns strong non-trivial team composition.
For fixed policy agents, we find that BERTeam learns the correct optimal team.
In the coevolution case, we find that BERTeam learns to form a balanced team of agents that performs competitively. 
Upon further inspection, we find that like its inspiration in the field of \gls{NLP}, BERTeam learns similarities between agent behaviors through initial embeddings.
This allows it to account for missing data by inferring the behavior of agents.
We also find that BERTeam outperforms \gls{MCAA}, an algorithm 
designed for team selection.


Overall, BERTeam is a strong team selection algorithm with roots inspired by \gls{NLP} text generation models. 
BERTeam's ability to learn similarities in agent behavior results in efficient training, and allows BERTeam to train alongside individual agent policies.

\subsection{Future Research}
\begin{itemize}

\item
We do not explore the capability to condition BERTeam's output on observations of the opponent.
In future research, we plan to show that teams generated through conditioning outperform teams generated with no information.

\item 
We test only size 2 teams to easily analyze the output of BERTeam. 
We plan to test larger teams in future research.

\item 
We can change our weighting and training so that BERTeam will converge on a Nash Equilibrium, assuming certain properties of the game outcomes (Appendix \ref{nashpendix}). 
We do not make these changes because they may be incompatible with \gls{MLM} training. 
We plan to explore this in future research.

\item 
BERTeam is applicable to games with more than two teams. 
In future experiments, it would be interesting to evaluate the performance of BERTeam on multi-team games.

\item
For coevolution, we only consider a basic evolutionary algorithm  with reproduction through cloning, as in \cite{coevdeeprl}. 
However, there is a vast literature of evolutionary algorithm variants that could replace this.
It would be interesting to explore which are most compatible with our training scheme.

\item
We do not focus on optimizing hyperparameters in our algorithms or their interactions. 
It would be interesting to optimize these across a wide variety of problem instances, and inspect their relation with aspects of each instance.

\commentout{
\item 
We focused our research on Pyquaticus, as this was the motivation for creating the BERTeam algorithm.
However, our algorithm is applicable to any multiagent adversarial team game.
It would be useful to verify the performance of BERTeam in future research with various problem instances.
}

\end{itemize}



\bibliographystyle{ACM-Reference-Format} 
\bibliography{refs}


\begin{thebibliography}{50}


\ifx \showCODEN    \undefined \def \showCODEN     #1{\unskip}     \fi
\ifx \showDOI      \undefined \def \showDOI       #1{#1}\fi
\ifx \showISBNx    \undefined \def \showISBNx     #1{\unskip}     \fi
\ifx \showISBNxiii \undefined \def \showISBNxiii  #1{\unskip}     \fi
\ifx \showISSN     \undefined \def \showISSN      #1{\unskip}     \fi
\ifx \showLCCN     \undefined \def \showLCCN      #1{\unskip}     \fi
\ifx \shownote     \undefined \def \shownote      #1{#1}          \fi
\ifx \showarticletitle \undefined \def \showarticletitle #1{#1}   \fi
\ifx \showURL      \undefined \def \showURL       {\relax}        \fi
\providecommand\bibfield[2]{#2}
\providecommand\bibinfo[2]{#2}
\providecommand\natexlab[1]{#1}
\providecommand\showeprint[2][]{arXiv:#2}

\bibitem[\protect\citeauthoryear{Bahdanau et~al\mbox{.}}{Bahdanau et~al\mbox{.}}{2016}]%
        {encdec1}
\bibfield{author}{\bibinfo{person}{Dzmitry Bahdanau} {et~al\mbox{.}}} \bibinfo{year}{2016}\natexlab{}.
\newblock \bibinfo{title}{{Neural Machine Translation by Jointly Learning to Align and Translate}}.
\newblock
\newblock
\showeprint[arxiv]{1409.0473}~[cs.CL]


\bibitem[\protect\citeauthoryear{Beason et~al\mbox{.}}{Beason et~al\mbox{.}}{2024}]%
        {pyquaticus}
\bibfield{author}{\bibinfo{person}{Jordan Beason} {et~al\mbox{.}}} \bibinfo{year}{2024}\natexlab{}.
\newblock \bibinfo{title}{{Evaluating Collaborative Autonomy in Opposed Environments using Maritime Capture-the-Flag Competitions}}.
\newblock
\newblock
\showeprint[arxiv]{2404.17038}~[cs.RO]


\bibitem[\protect\citeauthoryear{Beltagy et~al\mbox{.}}{Beltagy et~al\mbox{.}}{2020}]%
        {longformer}
\bibfield{author}{\bibinfo{person}{Iz Beltagy} {et~al\mbox{.}}} \bibinfo{year}{2020}\natexlab{}.
\newblock \bibinfo{title}{{Longformer: The Long-Document Transformer}}.
\newblock
\newblock
\showeprint[arxiv]{2004.05150}~[cs.CL]


\bibitem[\protect\citeauthoryear{Berger}{Berger}{2007}]%
        {fictitiousplay}
\bibfield{author}{\bibinfo{person}{Ulrich Berger}.} \bibinfo{year}{2007}\natexlab{}.
\newblock \showarticletitle{{Brown's original fictitious play}}.
\newblock \bibinfo{journal}{\emph{Journal of Economic Theory}} \bibinfo{volume}{135}, \bibinfo{number}{1} (\bibinfo{year}{2007}), \bibinfo{pages}{572--578}.
\newblock
\showISSN{0022-0531}


\bibitem[\protect\citeauthoryear{Beynier et~al\mbox{.}}{Beynier et~al\mbox{.}}{2013}]%
        {decmdp}
\bibfield{author}{\bibinfo{person}{Aur\'{e}lie Beynier} {et~al\mbox{.}}} \bibinfo{year}{2013}\natexlab{}.
\newblock \bibinfo{booktitle}{\emph{{DEC-MDP/POMDP}}}.
\newblock \bibinfo{publisher}{John Wiley \& Sons, Ltd}, Chapter~9, \bibinfo{pages}{277--318}.
\newblock
\showISBNx{9781118557426}


\bibitem[\protect\citeauthoryear{Brockman et~al\mbox{.}}{Brockman et~al\mbox{.}}{2016}]%
        {openaigym}
\bibfield{author}{\bibinfo{person}{Greg Brockman} {et~al\mbox{.}}} \bibinfo{year}{2016}\natexlab{}.
\newblock \bibinfo{title}{{OpenAI Gym}}.
\newblock
\newblock
\showeprint[arxiv]{1606.01540}~[cs.LG]


\bibitem[\protect\citeauthoryear{Brown}{Brown}{1951}]%
        {ogfictitiousplay}
\bibfield{author}{\bibinfo{person}{George Brown}.} \bibinfo{year}{1951}\natexlab{}.
\newblock \showarticletitle{{Iterative Solution of Games by Fictitious Play}}.
\newblock In \bibinfo{booktitle}{\emph{Activity Analysis of Production and Allocation}}, \bibfield{editor}{\bibinfo{person}{T.~C. Koopmans}} (Ed.). \bibinfo{publisher}{Wiley}.
\newblock


\bibitem[\protect\citeauthoryear{Brown et~al\mbox{.}}{Brown et~al\mbox{.}}{2020}]%
        {generation}
\bibfield{author}{\bibinfo{person}{Tom Brown} {et~al\mbox{.}}} \bibinfo{year}{2020}\natexlab{}.
\newblock \showarticletitle{{Language models are few-shot learners}}. In \bibinfo{booktitle}{\emph{Proceedings of the 34th International Conference on Neural Information Processing Systems}} \emph{(\bibinfo{series}{NIPS '20})}. \bibinfo{publisher}{Curran Associates Inc.}, Article \bibinfo{articleno}{159}, \bibinfo{numpages}{25}~pages.
\newblock
\showISBNx{9781713829546}


\bibitem[\protect\citeauthoryear{Chang et~al\mbox{.}}{Chang et~al\mbox{.}}{2020}]%
        {classification1}
\bibfield{author}{\bibinfo{person}{Wei-Cheng Chang} {et~al\mbox{.}}} \bibinfo{year}{2020}\natexlab{}.
\newblock \showarticletitle{{Taming Pretrained Transformers for Extreme Multi-label Text Classification}}. In \bibinfo{booktitle}{\emph{Proceedings of the 26th ACM SIGKDD International Conference on Knowledge Discovery \& Data Mining}} \emph{(\bibinfo{series}{KDD '20})}. \bibinfo{publisher}{Association for Computing Machinery}, \bibinfo{pages}{3163–3171}.
\newblock
\showISBNx{9781450379984}


\bibitem[\protect\citeauthoryear{Chen et~al\mbox{.}}{Chen et~al\mbox{.}}{2017}]%
        {reachavoid2}
\bibfield{author}{\bibinfo{person}{Mo Chen} {et~al\mbox{.}}} \bibinfo{year}{2017}\natexlab{}.
\newblock \showarticletitle{{Multiplayer Reach-Avoid Games via Pairwise Outcomes}}.
\newblock \bibinfo{journal}{\emph{IEEE Trans. Automat. Control}} \bibinfo{volume}{62}, \bibinfo{number}{3} (\bibinfo{year}{2017}), \bibinfo{pages}{1451--1457}.
\newblock


\bibitem[\protect\citeauthoryear{Cho et~al\mbox{.}}{Cho et~al\mbox{.}}{2014}]%
        {encdec2}
\bibfield{author}{\bibinfo{person}{Kyunghyun Cho} {et~al\mbox{.}}} \bibinfo{year}{2014}\natexlab{}.
\newblock \showarticletitle{{Learning Phrase Representations using RNN Encoder-Decoder for Statistical Machine Translation}}. In \bibinfo{booktitle}{\emph{Proceedings of the 2014 Conference on Empirical Methods in Natural Language Processing ({EMNLP})}}. \bibinfo{publisher}{Association for Computational Linguistics}, \bibinfo{pages}{1724--1734}.
\newblock


\bibitem[\protect\citeauthoryear{Cotton et~al\mbox{.}}{Cotton et~al\mbox{.}}{2020}]%
        {coevdeeprl}
\bibfield{author}{\bibinfo{person}{David Cotton} {et~al\mbox{.}}} \bibinfo{year}{2020}\natexlab{}.
\newblock \showarticletitle{{Coevolutionary Deep Reinforcement Learning}}. In \bibinfo{booktitle}{\emph{2020 IEEE Symposium Series on Computational Intelligence (SSCI)}}. \bibinfo{publisher}{Institute of Electrical and Electronics Engineers}, \bibinfo{pages}{2600--2607}.
\newblock


\bibitem[\protect\citeauthoryear{Devlin et~al\mbox{.}}{Devlin et~al\mbox{.}}{2019}]%
        {bert}
\bibfield{author}{\bibinfo{person}{Jacob Devlin} {et~al\mbox{.}}} \bibinfo{year}{2019}\natexlab{}.
\newblock \showarticletitle{{BERT: Pre-training of Deep Bidirectional Transformers for Language Understanding}}. In \bibinfo{booktitle}{\emph{Proceedings of NAACL-HLT}}, Vol.~\bibinfo{volume}{1}. \bibinfo{publisher}{Association for Computational Linguistics}, \bibinfo{pages}{2}.
\newblock


\bibitem[\protect\citeauthoryear{Dixit et~al\mbox{.}}{Dixit et~al\mbox{.}}{2022}]%
        {Dixit22}
\bibfield{author}{\bibinfo{person}{Gaurav Dixit} {et~al\mbox{.}}} \bibinfo{year}{2022}\natexlab{}.
\newblock \showarticletitle{{Diversifying behaviors for learning in asymmetric multiagent systems}}. In \bibinfo{booktitle}{\emph{Proceedings of the Genetic and Evolutionary Computation Conference}} \emph{(\bibinfo{series}{GECCO '22})}. \bibinfo{publisher}{Association for Computing Machinery}, \bibinfo{pages}{350–358}.
\newblock
\showISBNx{9781450392372}


\bibitem[\protect\citeauthoryear{Elo}{Elo}{1978}]%
        {elo}
\bibfield{author}{\bibinfo{person}{Arpad Elo}.} \bibinfo{year}{1978}\natexlab{}.
\newblock \bibinfo{booktitle}{\emph{{The Rating of Chessplayers, Past and Present}}}.
\newblock \bibinfo{publisher}{Arco Pub.}
\newblock
\showISBNx{0668047216 9780668047210}


\bibitem[\protect\citeauthoryear{Fedus, Goodfellow, and Dai}{Fedus et~al\mbox{.}}{2018}]%
        {bertrobust}
\bibfield{author}{\bibinfo{person}{William Fedus}, \bibinfo{person}{Ian~J. Goodfellow}, {and} \bibinfo{person}{Andrew~M. Dai}.} \bibinfo{year}{2018}\natexlab{}.
\newblock \showarticletitle{{MaskGAN: Better Text Generation via Filling in the {\_}\_\_\_\_\_\_}}. In \bibinfo{booktitle}{\emph{6th International Conference on Learning Representations, {ICLR} 2018, Vancouver, BC, Canada, April 30 - May 3, 2018, Conference Track Proceedings}}. \bibinfo{publisher}{OpenReview.net}.
\newblock


\bibitem[\protect\citeauthoryear{Flajolet and Sedgewick}{Flajolet and Sedgewick}{2013}]%
        {anal}
\bibfield{author}{\bibinfo{person}{Philippe Flajolet} {and} \bibinfo{person}{Robert Sedgewick}.} \bibinfo{year}{2013}\natexlab{}.
\newblock \bibinfo{booktitle}{\emph{{Analytic Combinatorics}}}.
\newblock \bibinfo{publisher}{Cambridge University Press}.
\newblock


\bibitem[\protect\citeauthoryear{Fr{\'i}as-Mart{\'i}nez and Sklar}{Fr{\'i}as-Mart{\'i}nez and Sklar}{2004}]%
        {robosoccer}
\bibfield{author}{\bibinfo{person}{Vanessa Fr{\'i}as-Mart{\'i}nez} {and} \bibinfo{person}{Elizabeth Sklar}.} \bibinfo{year}{2004}\natexlab{}.
\newblock \showarticletitle{{A team-based co-evolutionary approach to multi agent learning}}. In \bibinfo{booktitle}{\emph{Proceedings of the 2004 AAMAS Workshop on Learning and Evolution in Agent Based Systems}}. Citeseer, \bibinfo{publisher}{Autonomous Agents and Multiagent Systems}.
\newblock


\bibitem[\protect\citeauthoryear{Garcia et~al\mbox{.}}{Garcia et~al\mbox{.}}{2020}]%
        {reachavoid1}
\bibfield{author}{\bibinfo{person}{Eloy Garcia} {et~al\mbox{.}}} \bibinfo{year}{2020}\natexlab{}.
\newblock \showarticletitle{{Optimal Strategies for a Class of Multi-Player Reach-Avoid Differential Games in 3D Space}}.
\newblock \bibinfo{journal}{\emph{IEEE Robotics and Automation Letters}} \bibinfo{volume}{5}, \bibinfo{number}{3} (\bibinfo{year}{2020}), \bibinfo{pages}{4257--4264}.
\newblock


\bibitem[\protect\citeauthoryear{Hansen and Hurwitz}{Hansen and Hurwitz}{1943}]%
        {likelihood}
\bibfield{author}{\bibinfo{person}{Morris~H. Hansen} {and} \bibinfo{person}{William~N. Hurwitz}.} \bibinfo{year}{1943}\natexlab{}.
\newblock \showarticletitle{{On the Theory of Sampling from Finite Populations}}.
\newblock \bibinfo{journal}{\emph{The Annals of Mathematical Statistics}} \bibinfo{volume}{14}, \bibinfo{number}{4} (\bibinfo{year}{1943}), \bibinfo{pages}{333--362}.
\newblock
\showISSN{00034851}


\bibitem[\protect\citeauthoryear{Heinrich and Silver}{Heinrich and Silver}{2016}]%
        {selfplay}
\bibfield{author}{\bibinfo{person}{Johannes Heinrich} {and} \bibinfo{person}{David Silver}.} \bibinfo{year}{2016}\natexlab{}.
\newblock \bibinfo{title}{{Deep Reinforcement Learning from Self-Play in Imperfect-Information Games}}.
\newblock
\newblock
\showeprint[arxiv]{1603.01121}~[cs.LG]


\bibitem[\protect\citeauthoryear{Jaderberg et~al\mbox{.}}{Jaderberg et~al\mbox{.}}{2019}]%
        {jaderberg2019human}
\bibfield{author}{\bibinfo{person}{Max Jaderberg} {et~al\mbox{.}}} \bibinfo{year}{2019}\natexlab{}.
\newblock \showarticletitle{{Human-level performance in 3D multiplayer games with population-based reinforcement learning}}.
\newblock \bibinfo{journal}{\emph{Science}} \bibinfo{volume}{364}, \bibinfo{number}{6443} (\bibinfo{year}{2019}), \bibinfo{pages}{859--865}.
\newblock


\bibitem[\protect\citeauthoryear{Junczys-Dowmunt}{Junczys-Dowmunt}{2019}]%
        {translation2}
\bibfield{author}{\bibinfo{person}{Marcin Junczys-Dowmunt}.} \bibinfo{year}{2019}\natexlab{}.
\newblock \showarticletitle{{Microsoft Translator at WMT 2019: Towards Large-Scale Document-Level Neural Machine Translation}}. In \bibinfo{booktitle}{\emph{Proceedings of the Fourth Conference on Machine Translation (Volume 2: Shared Task Papers, Day 1)}}. \bibinfo{publisher}{Association for Computational Linguistics}, \bibinfo{pages}{225--233}.
\newblock


\bibitem[\protect\citeauthoryear{Kitano et~al\mbox{.}}{Kitano et~al\mbox{.}}{1997}]%
        {robocup}
\bibfield{author}{\bibinfo{person}{Hiroaki Kitano} {et~al\mbox{.}}} \bibinfo{year}{1997}\natexlab{}.
\newblock \showarticletitle{{The RoboCup synthetic agent challenge 97}}. In \bibinfo{booktitle}{\emph{Proceedings of the 15th International Joint Conference on Artifical Intelligence - Volume 1}} \emph{(\bibinfo{series}{IJCAI'97})}. \bibinfo{publisher}{Morgan Kaufmann Publishers Inc.}, \bibinfo{pages}{24–29}.
\newblock
\showISBNx{15558604804}


\bibitem[\protect\citeauthoryear{Klijn and Eiben}{Klijn and Eiben}{2021}]%
        {coevdeeprl2}
\bibfield{author}{\bibinfo{person}{Daan Klijn} {and} \bibinfo{person}{A.~E. Eiben}.} \bibinfo{year}{2021}\natexlab{}.
\newblock \showarticletitle{{A coevolutionary approach to deep multi-agent reinforcement learning}}. In \bibinfo{booktitle}{\emph{Proceedings of the Genetic and Evolutionary Computation Conference Companion}} \emph{(\bibinfo{series}{GECCO '21})}. \bibinfo{publisher}{Association for Computing Machinery}, \bibinfo{pages}{283–284}.
\newblock
\showISBNx{9781450383516}


\bibitem[\protect\citeauthoryear{Lin et~al\mbox{.}}{Lin et~al\mbox{.}}{2023}]%
        {lin2023tizero}
\bibfield{author}{\bibinfo{person}{Fanqi Lin} {et~al\mbox{.}}} \bibinfo{year}{2023}\natexlab{}.
\newblock \showarticletitle{{TiZero: Mastering Multi-Agent Football with Curriculum Learning and Self-Play}}. In \bibinfo{booktitle}{\emph{Proceedings of the 2023 International Conference on Autonomous Agents and Multiagent Systems}} \emph{(\bibinfo{series}{AAMAS '23})}. \bibinfo{publisher}{International Foundation for Autonomous Agents and Multiagent Systems}, \bibinfo{pages}{67–76}.
\newblock
\showISBNx{9781450394321}


\bibitem[\protect\citeauthoryear{Liu et~al\mbox{.}}{Liu et~al\mbox{.}}{2020}]%
        {translation1}
\bibfield{author}{\bibinfo{person}{Xiaodong Liu} {et~al\mbox{.}}} \bibinfo{year}{2020}\natexlab{}.
\newblock \bibinfo{title}{{Very Deep Transformers for Neural Machine Translation}}.
\newblock
\newblock
\showeprint[arxiv]{2008.07772}~[cs.CL]


\bibitem[\protect\citeauthoryear{McAleer et~al\mbox{.}}{McAleer et~al\mbox{.}}{2023}]%
        {mcaleer2023team}
\bibfield{author}{\bibinfo{person}{Stephen McAleer} {et~al\mbox{.}}} \bibinfo{year}{2023}\natexlab{}.
\newblock \showarticletitle{{Team-PSRO for Learning Approximate TMECor in Large Team Games via Cooperative Reinforcement Learning}}. In \bibinfo{booktitle}{\emph{Advances in Neural Information Processing Systems}}, \bibfield{editor}{\bibinfo{person}{A.~Oh}, \bibinfo{person}{T.~Naumann}, \bibinfo{person}{A.~Globerson}, \bibinfo{person}{K.~Saenko}, \bibinfo{person}{M.~Hardt}, {and} \bibinfo{person}{S.~Levine}} (Eds.), Vol.~\bibinfo{volume}{36}. \bibinfo{publisher}{Curran Associates, Inc.}, \bibinfo{pages}{45402--45418}.
\newblock


\bibitem[\protect\citeauthoryear{Mikolov et~al\mbox{.}}{Mikolov et~al\mbox{.}}{2013}]%
        {word2vec}
\bibfield{author}{\bibinfo{person}{Tomas Mikolov} {et~al\mbox{.}}} \bibinfo{year}{2013}\natexlab{}.
\newblock \bibinfo{title}{{Efficient estimation of word representations in vector space}}.
\newblock
\newblock
\showeprint[arxiv]{1301.3781}~[cs.CL]


\bibitem[\protect\citeauthoryear{Monderer and Shapley}{Monderer and Shapley}{1996a}]%
        {potential}
\bibfield{author}{\bibinfo{person}{Dov Monderer} {and} \bibinfo{person}{Lloyd Shapley}.} \bibinfo{year}{1996}\natexlab{a}.
\newblock \showarticletitle{{Fictitious Play Property for Games with Identical Interests}}.
\newblock \bibinfo{journal}{\emph{Journal of Economic Theory}} \bibinfo{volume}{68}, \bibinfo{number}{1} (\bibinfo{year}{1996}), \bibinfo{pages}{258--265}.
\newblock
\showISSN{0022-0531}


\bibitem[\protect\citeauthoryear{Monderer and Shapley}{Monderer and Shapley}{1996b}]%
        {potential2}
\bibfield{author}{\bibinfo{person}{Dov Monderer} {and} \bibinfo{person}{Lloyd~S. Shapley}.} \bibinfo{year}{1996}\natexlab{b}.
\newblock \showarticletitle{{Potential Games}}.
\newblock \bibinfo{journal}{\emph{Games and Economic Behavior}} \bibinfo{volume}{14}, \bibinfo{number}{1} (\bibinfo{year}{1996}), \bibinfo{pages}{124--143}.
\newblock
\showISSN{0899-8256}


\bibitem[\protect\citeauthoryear{Mouret and Clune}{Mouret and Clune}{2015}]%
        {mapelite}
\bibfield{author}{\bibinfo{person}{Jean-Baptiste Mouret} {and} \bibinfo{person}{Jeff Clune}.} \bibinfo{year}{2015}\natexlab{}.
\newblock \bibinfo{title}{{Illuminating search spaces by mapping elites}}.
\newblock
\newblock
\showeprint[arxiv]{1504.04909}~[cs.AI]


\bibitem[\protect\citeauthoryear{Novitzky et~al\mbox{.}}{Novitzky et~al\mbox{.}}{2019}]%
        {aquaticus}
\bibfield{author}{\bibinfo{person}{Michael Novitzky} {et~al\mbox{.}}} \bibinfo{year}{2019}\natexlab{}.
\newblock \showarticletitle{{Aquaticus: Publicly Available Datasets from a Marine Human-Robot Teaming Testbed}}. In \bibinfo{booktitle}{\emph{2019 14th ACM/IEEE International Conference on Human-Robot Interaction (HRI)}}. \bibinfo{publisher}{Institute of Electrical and Electronics Engineers}, \bibinfo{pages}{392--400}.
\newblock


\bibitem[\protect\citeauthoryear{Raffin et~al\mbox{.}}{Raffin et~al\mbox{.}}{2021}]%
        {stable-baselines3}
\bibfield{author}{\bibinfo{person}{Antonin Raffin} {et~al\mbox{.}}} \bibinfo{year}{2021}\natexlab{}.
\newblock \showarticletitle{{Stable-Baselines3: Reliable Reinforcement Learning Implementations}}.
\newblock \bibinfo{journal}{\emph{Journal of Machine Learning Research}} \bibinfo{volume}{22}, \bibinfo{number}{268} (\bibinfo{year}{2021}), \bibinfo{pages}{1--8}.
\newblock


\bibitem[\protect\citeauthoryear{Rajbhandari}{Rajbhandari}{2024a}]%
        {BERTeam}
\bibfield{author}{\bibinfo{person}{Pranav Rajbhandari}.} \bibinfo{year}{2024}\natexlab{a}.
\newblock \bibinfo{title}{{BERTeam}}.
\newblock \bibinfo{howpublished}{\url{https://github.com/pranavraj575/BERTeam}}.
\newblock


\bibitem[\protect\citeauthoryear{Rajbhandari}{Rajbhandari}{2024b}]%
        {coevolver}
\bibfield{author}{\bibinfo{person}{Pranav Rajbhandari}.} \bibinfo{year}{2024}\natexlab{b}.
\newblock \bibinfo{title}{{Transformer based Coevolver}}.
\newblock \bibinfo{howpublished}{\url{https://github.com/pranavraj575/coevolution}}.
\newblock


\bibitem[\protect\citeauthoryear{Rajbhandari}{Rajbhandari}{2024c}]%
        {unstable_baselines3}
\bibfield{author}{\bibinfo{person}{Pranav Rajbhandari}.} \bibinfo{year}{2024}\natexlab{c}.
\newblock \bibinfo{title}{{unstable\_baselines3}}.
\newblock \bibinfo{howpublished}{\url{https://github.com/pranavraj575/unstable_baselines3}}.
\newblock


\bibitem[\protect\citeauthoryear{Robinson}{Robinson}{1951}]%
        {zerosum}
\bibfield{author}{\bibinfo{person}{Julia Robinson}.} \bibinfo{year}{1951}\natexlab{}.
\newblock \showarticletitle{{An Iterative Method of Solving a Game}}.
\newblock \bibinfo{journal}{\emph{Annals of Mathematics}} \bibinfo{volume}{54}, \bibinfo{number}{2} (\bibinfo{year}{1951}), \bibinfo{pages}{296--301}.
\newblock
\showISSN{0003486X, 19398980}


\bibitem[\protect\citeauthoryear{Schulman et~al\mbox{.}}{Schulman et~al\mbox{.}}{2017}]%
        {ppo}
\bibfield{author}{\bibinfo{person}{John Schulman} {et~al\mbox{.}}} \bibinfo{year}{2017}\natexlab{}.
\newblock \bibinfo{title}{{Proximal Policy Optimization Algorithms}}.
\newblock
\newblock
\showeprint[arxiv]{1707.06347}~[cs.LG]


\bibitem[\protect\citeauthoryear{Shapley}{Shapley}{1953}]%
        {shapley1953stochastic}
\bibfield{author}{\bibinfo{person}{Lloyd Shapley}.} \bibinfo{year}{1953}\natexlab{}.
\newblock \showarticletitle{{Stochastic games}}.
\newblock \bibinfo{journal}{\emph{Proceedings of the National Academy of Sciences}} \bibinfo{volume}{39}, \bibinfo{number}{10} (\bibinfo{year}{1953}), \bibinfo{pages}{1095--1100}.
\newblock


\bibitem[\protect\citeauthoryear{Shishika et~al\mbox{.}}{Shishika et~al\mbox{.}}{2019}]%
        {team_comp_domain_specific}
\bibfield{author}{\bibinfo{person}{Daigo Shishika} {et~al\mbox{.}}} \bibinfo{year}{2019}\natexlab{}.
\newblock \showarticletitle{{Team Composition for Perimeter Defense with Patrollers and Defenders}}. In \bibinfo{booktitle}{\emph{2019 IEEE 58th Conference on Decision and Control (CDC)}}. \bibinfo{publisher}{Institute of Electrical and Electronics Engineers}, \bibinfo{pages}{7325--7332}.
\newblock


\bibitem[\protect\citeauthoryear{Song et~al\mbox{.}}{Song et~al\mbox{.}}{2024}]%
        {googlefootball}
\bibfield{author}{\bibinfo{person}{Yan Song} {et~al\mbox{.}}} \bibinfo{year}{2024}\natexlab{}.
\newblock \showarticletitle{{Boosting Studies of Multi-Agent Reinforcement Learning on Google Research Football Environment: The Past, Present, and Future}}. In \bibinfo{booktitle}{\emph{Proceedings of the 23rd International Conference on Autonomous Agents and Multiagent Systems}} \emph{(\bibinfo{series}{AAMAS '24})}. \bibinfo{publisher}{International Foundation for Autonomous Agents and Multiagent Systems}, \bibinfo{pages}{1772–1781}.
\newblock
\showISBNx{9798400704864}


\bibitem[\protect\citeauthoryear{Sutskever et~al\mbox{.}}{Sutskever et~al\mbox{.}}{2014}]%
        {encdec3}
\bibfield{author}{\bibinfo{person}{Ilya Sutskever} {et~al\mbox{.}}} \bibinfo{year}{2014}\natexlab{}.
\newblock \showarticletitle{{Sequence to sequence learning with neural networks}}. In \bibinfo{booktitle}{\emph{Proceedings of the 27th International Conference on Neural Information Processing Systems - Volume 2}} \emph{(\bibinfo{series}{NIPS'14})}. \bibinfo{publisher}{MIT Press}, \bibinfo{pages}{3104–3112}.
\newblock


\bibitem[\protect\citeauthoryear{Taylor}{Taylor}{2016}]%
        {cloze}
\bibfield{author}{\bibinfo{person}{Wilson Taylor}.} \bibinfo{year}{2016}\natexlab{}.
\newblock \showarticletitle{{"Cloze Procedure": A New Tool For Measuring Readability}}. In \bibinfo{booktitle}{\emph{Journalism Quarterly}}. \bibinfo{publisher}{Sage Journals}, \bibinfo{pages}{415--433}.
\newblock


\bibitem[\protect\citeauthoryear{Terry et~al\mbox{.}}{Terry et~al\mbox{.}}{2021}]%
        {pettingzoo}
\bibfield{author}{\bibinfo{person}{J.~K. Terry} {et~al\mbox{.}}} \bibinfo{year}{2021}\natexlab{}.
\newblock \showarticletitle{{PettingZoo: Gym for Multi-Agent Reinforcement Learning}}. In \bibinfo{booktitle}{\emph{Advances in Neural Information Processing Systems}}, Vol.~\bibinfo{volume}{34}. \bibinfo{publisher}{Curran Associates, Inc.}, \bibinfo{pages}{15032--15043}.
\newblock


\bibitem[\protect\citeauthoryear{Vaswani et~al\mbox{.}}{Vaswani et~al\mbox{.}}{2017}]%
        {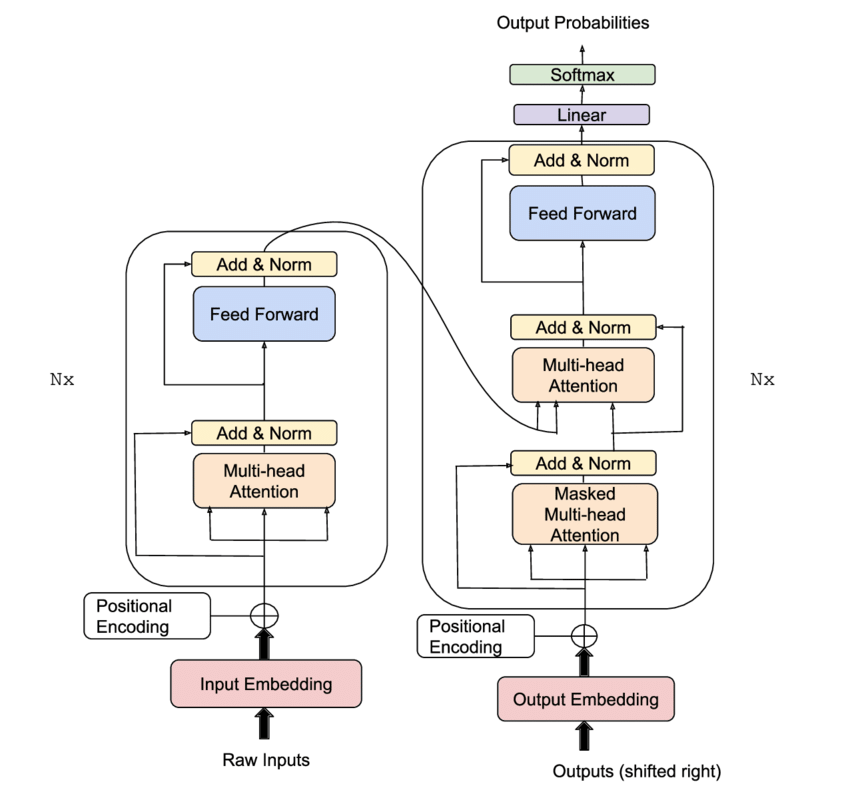}
\bibfield{author}{\bibinfo{person}{Ashish Vaswani} {et~al\mbox{.}}} \bibinfo{year}{2017}\natexlab{}.
\newblock \showarticletitle{{Attention is all you need}}. In \bibinfo{booktitle}{\emph{Proceedings of the 31st International Conference on Neural Information Processing Systems}} \emph{(\bibinfo{series}{NIPS'17})}. \bibinfo{publisher}{Curran Associates Inc.}, \bibinfo{pages}{6000–6010}.
\newblock
\showISBNx{9781510860964}


\bibitem[\protect\citeauthoryear{Yong and Miikkulainen}{Yong and Miikkulainen}{2001}]%
        {coop_co}
\bibfield{author}{\bibinfo{person}{Chern~Han Yong} {and} \bibinfo{person}{Risto Miikkulainen}.} \bibinfo{year}{2001}\natexlab{}.
\newblock \bibinfo{title}{{Cooperative Coevolution of Multi-Agent Systems}}.
\newblock
\newblock


\bibitem[\protect\citeauthoryear{Yong and Miikkulainen}{Yong and Miikkulainen}{2009}]%
        {coop_co2}
\bibfield{author}{\bibinfo{person}{Chern~Han Yong} {and} \bibinfo{person}{Risto Miikkulainen}.} \bibinfo{year}{2009}\natexlab{}.
\newblock \showarticletitle{{Coevolution of Role-Based Cooperation in Multiagent Systems}}.
\newblock \bibinfo{journal}{\emph{IEEE Transactions on Autonomous Mental Development}} \bibinfo{volume}{1}, \bibinfo{number}{3} (\bibinfo{year}{2009}), \bibinfo{pages}{170--186}.
\newblock


\bibitem[\protect\citeauthoryear{Zhao et~al\mbox{.}}{Zhao et~al\mbox{.}}{2021}]%
        {footballteammemberselection}
\bibfield{author}{\bibinfo{person}{Haoyu Zhao} {et~al\mbox{.}}} \bibinfo{year}{2021}\natexlab{}.
\newblock \showarticletitle{{Multi-Objective Optimization for Football Team Member Selection}}.
\newblock \bibinfo{journal}{\emph{IEEE Access}}  \bibinfo{volume}{9} (\bibinfo{year}{2021}), \bibinfo{pages}{90475--90487}.
\newblock


\bibitem[\protect\citeauthoryear{Zhao, Ju, and Hern{\'a}ndez-Orallo}{Zhao et~al\mbox{.}}{2024}]%
        {teamformationassesor}
\bibfield{author}{\bibinfo{person}{Yue Zhao}, \bibinfo{person}{Lushan Ju}, {and} \bibinfo{person}{Jos{\`e} Hern{\'a}ndez-Orallo}.} \bibinfo{year}{2024}\natexlab{}.
\newblock \showarticletitle{{Team formation through an assessor: choosing MARL agents in pursuit-evasion games}}.
\newblock \bibinfo{journal}{\emph{Complex {\&} Intelligent Systems}} \bibinfo{volume}{10}, \bibinfo{number}{3} (\bibinfo{year}{2024}), \bibinfo{pages}{3473--3492}.
\newblock
\showISSN{2198-6053}


\end{thebibliography}

\clearpage
\appendix
\section{Elo update equation}
\label{elopendix}

Consider a 1v1 game where outcomes for each player are non-negative and sum to 1. 
Elos are a method of assigning values to each agent in a population based on their ability in the game \cite{elo}.

If agents 1 and 2 with elos $f_1$ and $f_2$ play a game, we expect agent 1 to win with probability\footnote{Elos are scaled in chess by a factor of $\frac{400}{\log10}$, but ignoring this is cleaner} $Y_1:=\frac{1}{1+\exp(f_2-f_1)}$. 
When a game between agents 1 and 2 is sampled, the Elo of agent $i$ is updated with the agent's outcome $0\leq S_i\leq 1$ using the following equation:
\begin{equation}
    \label{eqn:elo_update}
    f_i'=f_i+c(S_i-Y_i).
\end{equation}
Note that if the outcome $S_i$ is larger (resp. smaller) than our expectation $Y_i$, we increase (resp. decrease) our Elo estimate. 
We set $c$ as the scale to determine the magnitude of the updates.

\section{Number of possible teams of size \textit{k}}
\label{number_of_possible_teams}
We will find the number of possible size $k$ teams with indistinguishable members from $n$ total policies. 

We first partition all possible teams of $k$ members based on their number of distinct policies $i$. 
In the case of $i$ policies, we must choose which of the $n$ policies to include ($\binom ni$ choices). 
We then assign an agent to each of the $k$ team members. 
Since we do not care about order, we must consider the number of ways to assign $k$ indistinguishable objects (members) into $i$ distinct bins (policies).
There are $\binom{k-1}{i-1}$ ways to do this by \lrquote{stars and bars} \cite{anal}. 
Thus, overall there are $\binom ni\binom{k-1}{i-1}$ team choices with $i$ distinct members. We sum this over all possible values of $i$ to obtain $\sum\limits_{i=1}^k\binom{n}{i}\binom{k-1}{i-1}$. 

If we do care about order (i.e. members are distinguishable), there are trivially $n^k$ ways to choose a sequence of $k$ agents from $n$ with repeats. 
Any intermediate order considerations must fall between these two extremes. 
In either case, with fixed $k$, there are $\Theta(n^k)$ possible choices of a team of size $k$ from $n$ total agents.
\section{Cosine similarity of random subset}
\label{cos_sim_proof}
A random subset of size $k$ chosen uniformly from a population of vectors will have the same average cosine similarity as the whole population.

For a proof, we may consider a fully connected graph where each vertex is one of $n$ vectors, and each edge joining two vertices is weighted by the cosine similarity of their vectors.
The above statement is equivalent to saying the expected average weight of edges in a random induced subgraph of size $k$ is the overall average edge weight. 

Consider taking the expectation across all possible $k$ subsets. Each edge weight will be counted $\binom{n-2}{k-2}$ times, as this is the number of $k$ subsets containing it (choose the other $k-2$ vertices from the $n-2$ remaining). 
When taking the edge average in each $k$ subset, we divide by $\binom k2$, and when taking the expectation across all $k$ subsets, we divide by $\binom nk$. Thus, an edge contributes 
$\frac{\binom {n-2}{k-2}}{\binom k2\binom nk}=
\binom n2^{-1}$ times its weight to the expectation, the same as it would in an average across all $\binom n2$ edges.

\section{Dataset Weighting/Sampling}
\label{weightpendix}
Consider the general case with $m$ teams in an adversarial game. 
We assume BERTeam has a current distribution for each team, and we would like to sample a dataset for the team in $i$th position.
We also assume we have a notion of a \lrquote{winning} team in each game. 
Our goal is that the occurrence of a team in the dataset is proportional to its win probability against opponents selected by BERTeam. 

For team $A$, denote this win probability $W(A)$. 
Let the set of all valid teams for the $i$th position be $\mathcal T_i$.

The na\"{i}ve approach is to simply sample uniformly from $\mathcal T_i$ and include teams that win against opponents sampled from BERTeam.
While this results in the correct distribution, this method will rarely find a successful team, as we assume BERTeam's choices are strong.

To increase the probability of finding a successful team, we may sample using BERTeam instead of uniformly from $\mathcal T_i$. 
This increases our success rate, but fails to generate the correct distribution.
In particular, the occurrence of team $A$ is proportional to $\mathbb P\{A\sim \text{BERTeam}\}\cdot W(A)$.
To fix this, we use inverse probability weighting \cite{likelihood},  weighting each inclusion of team $A$ by the inverse of its selection probability. 
This ensures that the \textit{weighted} occurrence of team $A$ in the dataset is $W(A)$. 

This additionally creates symmetry, since each team is sampled from BERTeam. 
Thus, we may consider building $m$ datasets and at each game, updating the datasets corresponding to winning teams. 
This increases the number of samples we get per game by a factor of $m$. 
In our experiments, where the players and opponents are symmetric, we keep one dataset and do this implicitly. 

\subsection{Relation to Nash Equilibria}
\label{nashpendix}

We analyze the general case where there are $m$ teams in a game, and the sets of valid teams are $\mathcal T_1,\dots,\mathcal T_m$.
The act of choosing teams to play multiagent adversarial matches suggests the structure of a normal form game with $m$ players (which we will call \lrquote{coaches} to distinguish from agents playing the multiagent game). 
The $i$th coach's available actions are teams in $\mathcal T_i$, and the utilities of a choice of $m$ teams are the expected outcomes of each team in the match.

The distribution of BERTeam defines a mixed strategy of a coach $i$ in this game (i.e. a distribution over all teams, an element of the simplex $\Delta(\mathcal T_i)$). 
Ideally, we would like the training to cause BERTeam to approach a Nash Equilibrium of the game. 
This is possible, under some assumptions.

We will define a loss function $\mathcal L'(p,T)$ for a distribution $p\in \Delta(\mathcal T_i)$ and sampled team $T\in \mathcal T_i$.
Let $\mathcal L'(p,T):=\|p-e_T\|_1$ be the L1 loss, where $e_T\in \Delta(\mathcal T_i)$ has a 1 only in the dimension corresponding to team $T$. 
For weighted dataset $S$, let $\mathcal L(p,S):=\sum\limits_{(T,w)\in S}w\mathcal L'(p,T)$.

Our first assumption is that there exists a BERTeam architecture and update scheme such that training on dataset $S$ will optimize $\mathcal L$ with respect to BERTeam's distribution $p\in \Delta(\mathcal T_i)$.

Now consider a dataset $S$ where the total weight of each team is its expected outcome against a distribution $q\in \prod\limits_{j\neq i}\Delta(\mathcal T_{j})$. 
We claim that an optimizer $p^*:=\underset{p\in\Delta(\mathcal T_i)}{\text{argmin}}\mathcal L(p,S)$ is a best response to $q$. 

Without loss of generality, we may assume $S$ contains exactly one copy of each team, and the weight of each team is exactly its expected outcome. 
Since elements in $p$ are nonnegative and sum to 1, we obtain 
$\mathcal L'(p,T)=
(1-p_T)+\sum\limits_{A\neq T}p_A=
2-2p_T
$, where $p_T$ is the element of $p$ on the dimension corresponding to $T\in\mathcal T_i$.
This implies $\mathcal L(p,S)=c-2\sum\limits_{(T,w)\in S}wp_T$. 
It is trivial to show this equation is minimized by a $p^*$ that puts all probability on teams with maximal weight. 
Since the weights correspond with each team's expected outcome against $q$, the possible values of $p^*$ is exactly the set of best responses to $q$ (distributions that put all probability on teams with maximal expected outcome).

This implies updating BERTeam to optimize loss $\mathcal L$ with respect to a dataset $S$ weighted in this way will result in an improving update to BERTeam's distribution $p$ based on empirical evidence of opponent strategies.
The same holds for any of the $m$ coaches.

This process is equivalent to \textit{fictitious play}, a process where players update their strategy based on empirical estimations of opponent strategies \cite{ogfictitiousplay,fictitiousplay}. 
There has been much research into for what classes of games fictitious play results in convergence to a Nash equilibrium (e.g. zero-sum games with finite strategies \cite{zerosum},
or potential games \cite{potential,potential2}). 
If our defined game happens to fall in any of these categories, this process will converge to a Nash equilibrium.
This is our second assumption.

To form a dataset $S$ such that the weighted occurrence of each team is its expected outcome against a distribution $q$, we can fill $S$ with teams in $\mathcal T_i$ weighed by sampled outcomes against $q$. 
In expectation, this will achieve our desired weighting.\footnote{
If we also must weight to account for sampling bias as in Appendix \ref{weightpendix}, we can simply multiply the inverse probability weight with the outcome weight to ensure both goals.
}

We implicitly assume that the behaviors of agents are fixed, so that games between two teams have constant expected outcome. 
This is certainly not true for configurations where we update agent policies, but can be \lrquote{true enough} if the policies have converged.

Our second assumption depends on the actual outcomes defined in the multiagent adversarial team game. 
However, even in situations where this assumption does not hold, updating BERTeam in this way will still result in fictitious play, a reasonable update strategy.
In the case of Pyquaticus, we can define outcome values to form a zero-sum game.

Our first assumption is untrue for \gls{MLM} training. 
Instead of a \lrquote{maximizing} update, \gls{MLM} training uses cross-entropy loss, encouraging the model to match a distribution.
While simply using L1 loss would create the correct optimization problem, there is little support in literature for using losses other than cross-entropy loss when working with transformer distributions.
Thus, even though the L1 loss has the correct optimal point, the \gls{MLM} training scheme may not be able to optimize it well.
Additionally, the output of a transformer is not a simple choice within $\Delta(\mathcal T_i)$.
With a team size larger than one, the method of choosing a team by iteratively picking members may also cause poorly behaved training.

\section{Experiment Parameters}
\label{parampendix}
\begin{table}[hb!]
    \centering
    \begin{tabular}{|c|c|c|}
    \hline
    \textbf{Parameter} & \textbf{Value} & \textbf{Justification} \\
    \hline
    Epochs & 3000 & Plot converged \\\hline
    Games per epoch & 25 & Consistency \\
    \hline
    \hline
    \multicolumn{3}{|c|}{\textit{BERTeam Transformer}}
    \\\hline
    Encoder/Decoder layers & 3/3 & $\frac12$ PyTorch default \\\hline
    Embedding dim & 128 & $\frac14$ PyTorch default \\\hline
    Feedforward dim & 512 & $\frac14$ PyTorch default \\\hline
    Attention heads & 4 & $\frac12$ PyTorch default \\\hline
    Dropout & 0.1 & PyTorch default \\\hline
    Train frequency & Every 10 epochs & Consistency \\\hline
    Batch/Minibatch size & 1024/256 &  \\
    \hline
    \hline
    \multicolumn{3}{|c|}{\textit{Input Embedding (Unused)}}\\\hline
    Network architecture & LSTM &  \\\hline
    Layers & 2 &  \\\hline
    Embedding dim & 128 & Same as BERTeam \\\hline
    Dropout & 0.1 & Same as BERTeam\\\hline
    \end{tabular}
    \caption{BERTeam experiment parameters (Section \ref{BERTeam_exp})}
    \label{berteam_params}
\end{table}

\begin{table}[hb!]
    \centering
    \begin{tabular}{|c|c|c|}
    \hline
    \textbf{Parameter} & \textbf{Value} & \textbf{Justification} \\
    \hline
    Epochs & 8000 &  \\\hline
    Games per epoch & 25 &  \\\hline
    \hline
    \multicolumn{3}{|c|}{\textit{Changes in BERTeam Parameters}}\\
    \hline
    Batch/Minibatch size & 512/64 & Lowered for speed \\
    \hline
    \hline
    \multicolumn{3}{|c|}{\textit{Coevolution}}\\\hline
    Population Size & 50 &  \\\hline
    
    Replacements & \multirow{2}{*}{1} & Drastic changes may \\
    per generation &  & destabilize BERTeam \\\hline
    
    Protection of & \multirow{2}{*}{500 epochs} & Decent policies require \\
    new agents &  & $\sim 500$ games of training \\\hline
    
    Elite agents & 3 & Protect best policies \\
    \hline
    \hline
    \multicolumn{3}{|c|}{\textit{Reinforcement Learning}}\\\hline
    \gls{RL} algorithm & PPO &  \\\hline
    Network hidden layers & $(64,64)$ & stable\_baselines3 default \\\hline
    \end{tabular}
    \caption{Coevolution experiment parameters (Section \ref{coevolution_exp})}
    \label{coevolution_params}
\end{table}

\begin{table}[hb!]
    \centering
    \begin{tabular}{|c|c|c|}
    \hline
    \textbf{Parameter} & \textbf{Value} & \textbf{Justification} 
    \\\hline
    Epochs & 4000 & Decreased for speed \\
    \hline
    Games per Epoch & 16 & Decreased for speed \\\hline
    Islands & 4 & Same as experiments in \cite{Dixit22} \\\hline
    Island Size & 15 & $4\cdot 15\approx 50$ \\\hline
    Elite Agents per Island & 1 & $1\cdot 4\approx 3$ \\
    \hline
    \hline
    \multicolumn{3}{|c|}{\textit{Island 0}}\\\hline
    \gls{RL} algorithm & PPO &  \\\hline
    Network hidden layers & $(64,64)$ & stable\_baselines3 default \\
    \hline
    \hline
    \multicolumn{3}{|c|}{\textit{Island 1}}\\\hline
    \gls{RL} algorithm & PPO &  \\\hline
    Network hidden layers & $(96,96)$ &  Slightly more complex\\\hline 
    \hline
    \multicolumn{3}{|c|}{\textit{Island 2}}\\\hline
    \gls{RL} algorithm & DQN &  \\\hline
    Network hidden layers & $(64,64)$ & stable\_baselines3 default \\
    \hline
    \hline
    \multicolumn{3}{|c|}{\textit{Island 3}}\\\hline
    \gls{RL} algorithm & DQN &  \\\hline
    Network hidden layers & $(96,96)$ & Slightly more complex \\\hline
    \end{tabular}
    \caption{Comparison experiment parameters (Section \ref{comparison_exp})}
    \label{commparison_params}
\end{table}

\end{document}